\documentclass{article}

\PassOptionsToPackage{round}{natbib}

\usepackage[main, final]{neurips_2026}

\usepackage[utf8]{inputenc} 
\usepackage[T1]{fontenc}    
\usepackage{hyperref}       
\usepackage{url}            
\usepackage{booktabs}       
\usepackage{amsfonts}       
\usepackage{nicefrac}       
\usepackage{microtype}      
\usepackage{xcolor}         

\usepackage{graphicx}
\usepackage{subcaption}
\usepackage{multirow}
\usepackage[normalem]{ulem}
\usepackage{xurl}
\usepackage{amsmath}
\usepackage{amssymb}
\usepackage{mathtools}
\usepackage{amsthm}

\usepackage{algorithm}
\usepackage{algorithmic}
\usepackage{float}  

\usepackage[capitalize,noabbrev]{cleveref}

\let\cite\citep
\theoremstyle{plain}

\theoremstyle{definition}

\theoremstyle{remark}

\usepackage[disable,textsize=tiny]{todonotes}

\title{COMET: Codebook-based Online-adaptive Multi-scale Embedding for Time-series Anomaly Detection}

\author{%
  Jinwoo Park \\
  Department of Industrial Engineering\\
  Seoul National University\\
  Seoul, Republic of Korea \\
  \texttt{jinwoo\_park@snu.ac.kr} \\
  \And
  Hyeongwon Kang \\
  Department of Industrial \& Management Engineering\\
  Korea University\\
  Seoul, Republic of Korea \\
  \texttt{hyeongwon\_kang@korea.ac.kr} \\
  \And
  Seung Hun Han \\
  LG CNS, Republic of Korea \\
  \texttt{seunghun.han@lgcns.com} \\
  \And
  Pilsung Kang\thanks{Corresponding author.} \\
  Department of Industrial Engineering\\
  Seoul National University\\
  Seoul, Republic of Korea \\
  \texttt{pilsung\_kang@snu.ac.kr} \\
}

\begin{document}

\maketitle

\begin{abstract}
    Multivariate time-series anomaly detection aims to identify abnormal temporal patterns in complex real-world systems. However, robust unsupervised time-series anomaly detection remains challenging because anomalies can manifest over different temporal ranges and emerge from interactions among variables, while distribution shifts can cause normal patterns to drift over time. To address these challenges, we propose Codebook-based Online-adaptive Multi-scale Embedding for Time-series anomaly detection (COMET), a unified framework with three tightly coupled components. First, Multi-scale Patch Encoder learns correlation-aware patch embeddings by combining variable-specific temporal dynamics with shared inter-variable structure across multiple patch scales. Second, Vector-Quantized Coreset quantizes these embeddings into representative normal prototypes, which are used to detect anomalies through a dual score combining quantization error and density-aware memory distance. Third, Online Codebook Adaptation leverages the same codebook structure to identify reliable normal samples from training-activated codebook entries, and then updates the prototypes through contrastive learning at inference time. Experiments on five benchmark datasets demonstrate that COMET achieves the best performance on 39 out of 45 evaluation metrics compared to seven baselines, showing robust performance across diverse evaluation protocols.
\end{abstract}

\section{Introduction}
\label{sec:Introduction}

Time-series anomaly detection has become a core analytical tool for ensuring reliable system operation, particularly as large-scale sensor systems are increasingly deployed across diverse industrial domains~\cite{wang2022detecting, zamanzadeh2024deep}. Because anomalies are rare and costly to label, unsupervised approaches are widely adopted: models learn normal patterns from training data and detect anomalies as deviations from these learned patterns~\cite{mejri2024unsupervised}. Following standard convention, we use the term \emph{unsupervised} to indicate that no explicit anomaly labels are used during training, while assuming that the training data are predominantly normal.

Unsupervised time-series anomaly detection still faces several challenges arising from the inherent properties of time series and the task-specific requirements of anomaly detection. First, effective embeddings should capture both temporal dependencies and multivariate correlations~\cite{li2021multivariate}. Although recent patch-based methods~\cite{NieNSK23} encode local temporal patterns, they often rely on a single fixed patch scale and channel-independent architectures. As a result, they may overlook inter-variable relationships and struggle to represent patterns that emerge across diverse temporal scales~\cite{park2025granularity, lu2022multi}. Second, many existing methods compute anomaly scores at the timestep level, limiting their ability to capture contextual and collective anomalies~\cite{choi2021deep, lai2023nominality}. While point anomalies can often be detected from timestep-wise deviations, contextual and collective anomalies are characterized by deviations that become evident only over broader temporal contexts. Third, real-world time series are inherently non-stationary~\cite{liu2023koopa, kim2021reversible}, which often induces distribution shifts between training and inference and thereby degrades generalization. Although Test-Time Adaptation (TTA) has been explored to address such shifts~\cite{kim2024model}, existing approaches typically rely on threshold-based filtering and lack explicit mechanisms for reliably identifying normal test samples. Consequently, anomalous patterns may be mistakenly assimilated into the normal representation during adaptation.

In this paper, we propose Codebook-based Online-adaptive Multi-scale Embedding for Time-series anomaly detection (COMET). COMET models temporal dependencies and multivariate correlations through a Multi-scale Patch Encoder, detects anomalies at the patch level by learning representative normal prototypes with a Vector-Quantized Coreset, and adapts to inference-time distribution shifts using Online Codebook Adaptation. The main contributions of this work are summarized as follows:
\begin{itemize}
\item We propose a Multi-scale Patch Encoder that captures both temporal dependencies and inter-variable correlations by combining variable-wise encoders with a shared core encoder across multiple patch scales.
\item We introduce a Vector-Quantized Coreset that learns representative prototypes of normal patterns and detects anomalies using a dual scoring scheme based on quantization error and memory distance.
\item We propose Online Codebook Adaptation, which mitigates distribution shifts during inference by identifying reliable normal test samples through codebook activation-based pseudo-labeling and adapting the representation with contrastive learning.
\item We conduct extensive experiments on five benchmark datasets and show that COMET achieves state-of-the-art performance on 39 out of 45 evaluation metrics without test-time adaptation, compared with seven baselines.
\end{itemize}

\section{Related work}
\label{sec:Related-work}

\subsection{Unsupervised multivariate time-series anomaly detection}
\label{subsec:Unsupervised-Multivariate-Time-Series-Anomaly-Detection}

\textbf{Reconstruction-based methods.} 
Reconstruction-based methods train models on normal data and detect anomalies using reconstruction error, under the assumption that anomalous patterns are difficult to reconstruct~\cite{zamanzadeh2024deep}. Early studies modeled temporal dependencies with LSTM-based autoencoders~\cite{LSTMAE} and their variational extensions~\cite{LSTMVAE}, while USAD~\cite{USAD} further amplified reconstruction errors through adversarial training. With the success of Transformers, attention-based approaches such as Anomaly Transformer~\cite{AnomalyTransformer} and VTT~\cite{VTT} have been proposed, and recent methods have further addressed non-stationarity~\cite{D3R} and frequency-aware multivariate patterns~\cite{CATCH}. Although these models can encode contextual information during reconstruction, their anomaly scores are typically computed at the timestep level. This can lead to under-detection when individual timesteps appear normal in isolation but collectively deviate from normal temporal patterns~\cite{lai2023nominality}. COMET differs from this paradigm by computing anomaly scores at the patch level, thereby preserving contextual structure directly in the scoring process.

\textbf{Representation-based methods.}
Representation-based methods detect anomalies using learned latent representations rather than raw input observations~\cite{zamanzadeh2024deep}. THOC~\cite{THOC} learns hierarchical temporal representations and applies one-class classification, DCdetector~\cite{DCdetector} combines dual-view attention with contrastive learning, and CARLA~\cite{CARLA} learns decision boundaries through self-supervised learning with synthetic anomaly injection. COMET also follows a representation-based perspective, but it models normal patterns using codebook entries learned through vector quantization. These entries serve as a compact coreset of normal prototypes, enabling distance-based anomaly scoring in the learned representation space.

\subsection{Vector quantization for time series}
\label{subsec:Vector-Quantization-for-Time-Series}

Vector Quantization (VQ) maps continuous representations to a finite set of discrete codebook entries. VQ-VAE~\cite{VQVAE} introduced this approach for learning discrete latent representations, and TimeVQVAE~\cite{TimeVQVAE} extended it to time series generation by modeling temporal dynamics in a quantized latent space. TimeVQVAE-AD~\cite{TimeVQVAEAD} further applied this framework to anomaly detection, leveraging masked generative modeling to identify abnormal segments based on reconstruction difficulty. In contrast, COMET employs VQ to learn representative prototypes of normal patterns and detects anomalies based on distances in the learned embedding space.

\subsection{Test-time adaptation for time series}
\label{subsec:Test-Time-Adaptation-for-Time-Series}

Test-Time Adaptation (TTA) adapts a trained model to incoming test data and is particularly relevant for time series, where non-stationarity can induce distribution shifts between training and inference. While several TTA methods have been proposed for time-series forecasting~\cite{TAFAS, PETSA, PROCEED}, TTA for anomaly detection remains relatively underexplored. M2N2~\cite{kim2024model} updates model parameters using samples classified as normal, but its reliance on threshold-based filtering makes adaptation sensitive to threshold selection and potential misclassification. In contrast, COMET identifies reliable normal samples through codebook entries activated during training, thereby reducing reliance on explicit threshold-based filtering during adaptation.

\section{Proposed method}
\label{sec:Proposed-Method}

\textbf{Problem Definition.}
Given a multivariate time series $\mathbf{X} \in \mathbb{R}^{L \times D}$ with sequence length $L$ and $D$ variables, our goal is to train a model on unlabeled, predominantly normal data and produce an anomaly score $S_t \in \mathbb{R}$ for each timestep $t$ during inference.

\textbf{Pipeline Overview.}
COMET is a unified framework whose components are designed to address the three challenges discussed in Section~\ref{sec:Introduction}. \emph{Multi-scale Patch Encoder} (Section~\ref{subsec:Multi-scale-Patch-Encoding}) generates patch embeddings that capture temporal dependencies and inter-variable correlations across multiple scales. Based on these embeddings, \emph{Vector-Quantized Coreset} (Section~\ref{subsec:Vector-Quantized-Coreset}) compresses normal patterns into discrete prototypes and computes patch-level anomaly scores, which are subsequently mapped to timestep-level scores while preserving contextual and collective information. During inference, \emph{Online Codebook Adaptation} (Section~\ref{subsec:Online-Codebook-Adaptation}) uses codebook entries activated during training to identify reliable normal samples without relying on explicit score thresholds, thereby mitigating distribution shifts while reducing the risk of adapting to misclassified anomalies. Figure~\ref{fig:figure1} illustrates the overall pipeline, and Algorithms~\ref{alg:training} and~\ref{alg:inference} in Appendix~\ref{appendix:algorithm} formalize the training and inference procedures.

\begin{figure}[t]
  \begin{center}
    \centerline{\includegraphics[width=\linewidth]{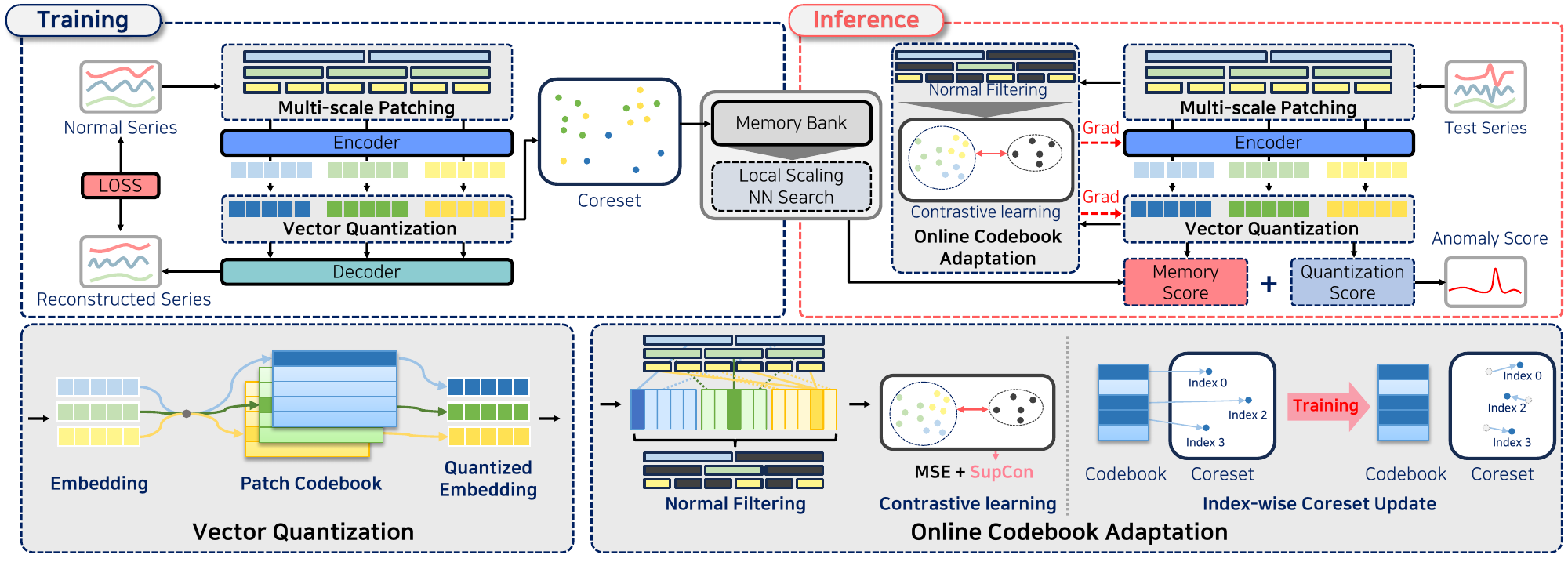}}
    \caption{Overall architecture of COMET. \textbf{(Top-left)} During training, input time series are processed via \emph{Multi-scale Patch Encoder} to capture temporal contexts at different resolutions. The resulting embeddings are mapped to a \emph{Vector-Quantized Coreset}, where activated codebook entries are stored in the memory bank. \textbf{(Top-right)} During inference, \emph{Online Codebook Adaptation} selectively updates the codebook by identifying reliable normal samples through pseudo-labeling and contrastive learning. Anomaly scores are computed by combining memory distance and quantization error. \textbf{(Bottom-left)} Detailed illustration of vector quantization, where continuous patch embeddings are assigned to their nearest codebook entries and represented by the corresponding quantized embeddings. \textbf{(Bottom-right)} Detailed illustration of \emph{Online Codebook Adaptation}, comprising normal filtering, contrastive learning, and index-wise coreset update. Reliable normal samples are selected based on training-activated codebook entries, and the corresponding codebook prototypes are updated to adapt to distribution shifts while reducing contamination from anomalous patterns.}
    \label{fig:figure1}
  \end{center}
  \vskip -0.3in
\end{figure}

\subsection{Multi-scale Patch Encoder}
\label{subsec:Multi-scale-Patch-Encoding}

To capture both multi-scale temporal patterns and inter-variable correlations, we propose Multi-scale Patch Encoder, which combines variable-wise encoders with a shared core encoder across multiple patch scales, inspired by SOFTS~\cite{SOFTS}.

\textbf{Multi-scale Patching.} Given an input time series $\mathbf{X} \in \mathbb{R}^{L \times D}$, we extract multi-scale patches using $K$ different patch sizes $\mathcal{P} = \{p_1, ..., p_K\}$ with corresponding strides $\mathcal{S} = \{s_1, ..., s_K\}$. At scale $k$, the $j$-th patch of variable $i$ is defined as
\begin{equation}
    \mathbf{P}_k^{(i,j)} = \mathbf{X}_{j \cdot s_k : j \cdot s_k + p_k, \, i} \in \mathbb{R}^{p_k},
\end{equation}
where $i \in \{1, ..., D\}$ and $j \in \{0, ..., N_k - 1\}$, and the number of patches is given by $N_k = \lceil (L - p_k) / s_k \rceil + 1$.

\textbf{Variable-wise Patch Encoder.} Since each variable in a multivariate time series exhibits distinct statistical characteristics and patterns, we apply an independent encoder $f_k^{(i)}: \mathbb{R}^{p_k} \rightarrow \mathbb{R}^{d/2}$ to each variable $i$:
\begin{equation}
    \mathbf{h}_k^{(i,j)}=f_k^{(i)}(\mathbf{P}_k^{(i,j)})=\mathbf{W}_k^{(i)}\mathbf{P}_k^{(i,j)}+\mathbf{b}_k^{(i)},
\end{equation}
where $\mathbf{W}_k^{(i)} \in \mathbb{R}^{(d/2) \times p_k}$ denotes learnable weights and $\mathbf{b}_k^{(i)} \in \mathbb{R}^{d/2}$ denotes a bias term.

\textbf{Core Encoder.} To address the limitation that variable-specific features alone cannot capture inter-variable correlations, we concatenate patches from all variables at each patch index $j$ and employ a shared encoder $f_k^{(c)}: \mathbb{R}^{D \cdot p_k} \rightarrow \mathbb{R}^{d_c}$ with core encoder output dimension $d_c$ to model correlations across variables:
\begin{equation}
\tilde{\mathbf{P}}_k^{(j)} = \left[ \mathbf{P}_k^{(1,j)}; \cdots; \mathbf{P}_k^{(D,j)} \right] \in \mathbb{R}^{D \cdot p_k},
\quad
\mathbf{h}_k^{(c,j)} = f_k^{(c)}(\tilde{\mathbf{P}}_k^{(j)})
= \mathbf{W}_k^{(c)} \tilde{\mathbf{P}}_k^{(j)} + \mathbf{b}_k^{(c)}.
\end{equation}

The core encoder shares parameters across all variables within the same scale, enabling it to learn inter-variable interactions.

\textbf{Feature Fusion.} To effectively integrate variable-specific features $\mathbf{h}_k^{(i,j)}$ with shared features $\mathbf{h}_k^{(c,j)}$, we concatenate them and apply a fusion layer:
\begin{equation}
    \mathbf{z}_{e,k}^{(i,j)} = \mathbf{W}_k^{(g)} \left[ \mathbf{h}_k^{(i,j)}; \mathbf{h}_k^{(c,j)} \right] + \mathbf{b}_k^{(g)},
\end{equation}
where $\mathbf{W}_k^{(g)} \in \mathbb{R}^{d \times ((d/2) + d_c)}$ and $d$ is the fused patch embedding dimension. Through the feature fusion process, we obtain a continuous embedding $\mathbf{Z}_{e,k} \in \mathbb{R}^{D \times N_k \times d}$ at scale $k$.

\subsection{Vector-Quantized Coreset}
\label{subsec:Vector-Quantized-Coreset}

To learn representative prototypes of normal patterns in a memory-efficient manner, we leverage Vector Quantization (VQ), which quantizes continuous embeddings into a finite set of codebook entries that serve as prototypes of normal patterns. The resulting memory bank uses at most $M$ entries per scale, and storing the quantization indices enables efficient index-based updates during Online Codebook Adaptation.

\textbf{Vector Quantization.} For each scale $k$, we define a learnable codebook $\mathbf{C}_k = \{\mathbf{c}_{k,m}\}_{m=1}^M \in \mathbb{R}^{M \times d}$. The continuous embedding $\mathbf{z}_{e,k}^{(i,j)}$ is then quantized to the nearest entry in the codebook as follows:
\begin{equation}
    q_k^{(i,j)} = \underset{m \in \{1, ..., M\}}{\arg\min} \left\| \mathbf{z}_{e,k}^{(i,j)} - \mathbf{c}_{k,m} \right\|_2^2,
    \quad
    \mathbf{z}_{q,k}^{(i,j)} = \mathbf{c}_{k,q_k^{(i,j)}},
\end{equation}
where $q_k^{(i,j)} \in \{1, ..., M\}$ denotes the quantization index, and $\mathbf{z}_{q,k}^{(i,j)} \in \mathbb{R}^d$ represents the quantized embedding.

\textbf{Training Objective.} To train the vector quantization module, we employ a codebook loss and a commitment loss:
\begin{equation}
    \mathcal{L}_{\text{cb}} = \left\| \mathbf{z}_{q,k}^{(i,j)} - \text{sg}[\mathbf{z}_{e,k}^{(i,j)}] \right\|_2^2,
    \quad
    \mathcal{L}_{\text{cm}} = \left\| \text{sg}[\mathbf{z}_{q,k}^{(i,j)}] - \mathbf{z}_{e,k}^{(i,j)} \right\|_2^2,
\end{equation}
where $\text{sg}[\cdot]$ denotes the stop-gradient operator. The codebook loss updates the codebook entries toward the encoder outputs, while the commitment loss encourages the encoder outputs to remain close to the selected codebook entries. The quantized embeddings are subsequently passed through a decoder $h_k(\cdot)$ to reconstruct the original patches, yielding the following reconstruction loss:
\begin{equation}
    \mathcal{L}_{\text{rec}} = \left\| \mathbf{P}_k^{(i,j)} - h_k(\mathbf{z}_{q,k}^{(i,j)}) \right\|_2^2.
\end{equation}
The overall training objective combines the reconstruction loss with the VQ losses as follows:
\begin{equation}
    \mathcal{L}_{\text{total}} = \mathcal{L}_{\text{rec}} + \alpha \mathcal{L}_{\text{cb}} + \beta \mathcal{L}_{\text{cm}},
\end{equation}
where $\alpha$ and $\beta$ are weighting coefficients for the respective VQ loss terms.

\textbf{Coreset Memory Bank.} After training, we construct per-scale coresets $\{\mathcal{M}_k\}_{k=1}^{K}$ by collecting the codebook entries activated by the training data at each scale:
\begin{equation}
\label{eq:M_k}
    \mathcal{M}_k = \left\{ \mathbf{c}_{k,m} \mid m \in \mathcal{A}_k \right\},
    \quad k=1,\ldots,K,
\end{equation}
where $\mathcal{A}_k \subseteq \{1, ..., M\}$ denotes the set of codebook indices activated at scale $k$ during training. For anomaly scoring, all scales share a single memory bank formed by the union of the per-scale coresets, $\mathcal{M}=\bigcup_{k=1}^{K}\mathcal{M}_k$.

\subsection{Anomaly scoring}
\label{subsec:Anomaly-Scoring}

At inference time, anomalies are detected using two complementary scores: (1) a memory score measuring distances to the memory bank and (2) a quantization score capturing quantization error.

\begin{figure}[t]
  \begin{center}
    \centerline{\includegraphics[width=\linewidth]{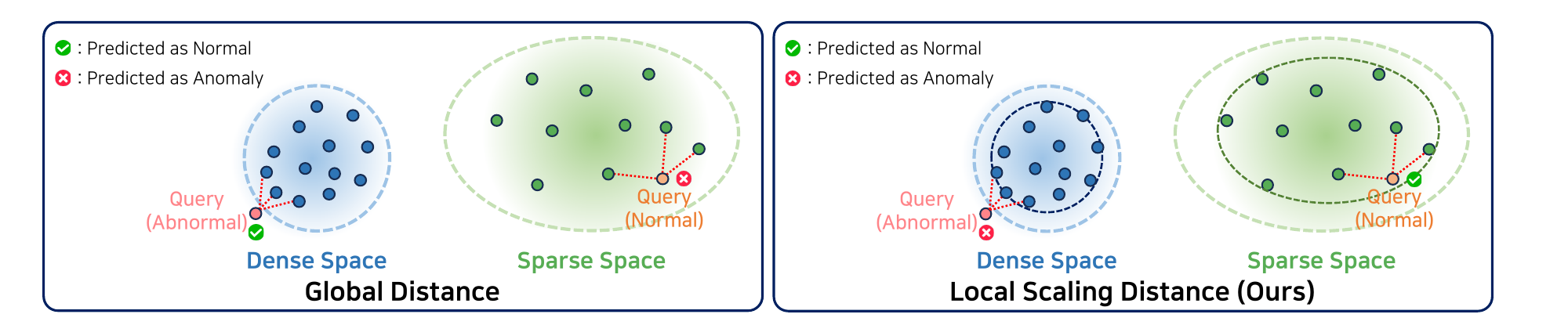}}
    \caption{Comparison between global distance and local scaling distance. \textbf{(Left)} Global distance fails to identify normal queries in sparse regions as it relies solely on absolute distance without considering local density. \textbf{(Right)} Local scaling distance accounts for local density, leading to more consistent classification of queries across regions with varying density.}
    \label{fig:figure2}
  \end{center}
  \vskip -0.3in
\end{figure}

\textbf{Memory Score with Local Scaling Distance.} The memory score measures the distance between a quantized embedding and its nearest normal prototypes in the memory bank. Since the memory bank is constructed from training-activated codebook entries, its sample density can be highly non-uniform, so using only a global distance may produce false positives in sparse normal regions and false negatives in dense regions, as illustrated in Figure~\ref{fig:figure2}. To address this, we introduce a local scaling distance that jointly accounts for distance and local density. Leveraging the fact that nearest-neighbor distances serve as a proxy for local neighborhood density~\cite{zelnik2004self}, we define a local scale $\sigma_x$ for both memory samples $\mathbf{m}_i \in \mathcal{M}$ and quantized query embeddings $\mathbf{z}_{q,k}$ as the median squared distance to their $n$ nearest neighbors within $\mathcal{M}$, where $\mathcal{N}_n(\cdot)$ denotes the neighbor set:
\begin{equation}
    \sigma_x = \mathrm{median}\left( \left\{ \left\| \mathbf{x} - \mathbf{m}_j \right\|_2^2 \;\middle|\; \mathbf{m}_j \in \mathcal{N}_{n}(\mathbf{x}) \right\} \right), \quad \mathbf{x} \in \{\mathbf{m}_i, \mathbf{z}_{q,k}\}.
\end{equation}
The local scaling distance normalizes the squared Euclidean distance between the quantized query embedding $\mathbf{z}_q$ and a neighbor $\mathbf{m}_j$ by the average of their local scales:
\begin{equation}
    d_{\text{local}}(\mathbf{z}_q, \mathbf{m}_j) = \frac{\| \mathbf{z}_q - \mathbf{m}_j \|_2^2}{(\sigma_{\mathbf{z}_q} + \sigma_{\mathbf{m}_j}) / 2 + \epsilon}.
\end{equation}
The proposed local scaling distance enables sensitivity to small deviations in dense regions while tolerating larger deviations in sparse regions, achieving density-adaptive anomaly detection. Finally, for each patch scale $k$, we compute the average local scaling distance over the $n$ nearest neighbors to obtain the patch-level memory score, which is then averaged over all scales:
\begin{equation}
S_{\text{mem},k}^{(i,j)}
=
\frac{1}{n}
\sum_{r=1}^{n}
d_{\text{local}}\!\left( \mathbf{z}_{q,k}^{(i,j)}, \mathbf{m}_{r}^{(k)} \right),
\qquad
S_{\text{mem}}^{(t)}
=
\frac{1}{K}
\sum\nolimits_{k=1}^{K}
\bar{S}_{\text{mem},k}^{(t)},
\end{equation}
where $\{\mathbf{m}_{r}^{(k)}\}_{r=1}^{n}$ are the $n$ nearest memory samples to $\mathbf{z}_{q,k}^{(i,j)}$ in $\mathcal{M}$, which may also include representations from other scales, and $\bar{S}_{\text{mem},k}^{(t)}$ denotes the score at scale $k$ for the $t$-th test batch, mapped to timesteps after deviation-based variable selection.

\textbf{Quantization Score.} 
The quantization score measures how well the encoder output $\mathbf{z}_e$ is represented by the codebook. Since the codebook is trained only on normal data, normal patterns are expected to yield low quantization errors, whereas anomalous patterns tend to produce larger errors due to the absence of similar codebook entries:
\begin{equation}
S_{\text{quant},k}^{(i,j)}
=
\left\|
\mathbf{z}_{e,k}^{(i,j)}
-
\mathbf{z}_{q,k}^{(i,j)}
\right\|_2,
\qquad
S_{\text{quant}}^{(t)}
=
\frac{1}{K}
\sum\nolimits_{k=1}^{K}
\bar{S}_{\text{quant},k}^{(t)}.
\end{equation}
As with the memory score, the quantization score is mapped to timesteps as $\bar{S}_{\text{quant},k}^{(t)}$.

\textbf{Deviation-based Variable Selection.}
In multivariate time series, aggregating scores across all variables can be unreliable due to redundant or noisy channels. We therefore perform temporal position-wise variable selection based on standardized temporal deviations, retaining stable variables while excluding unstable ones. Further details are provided in Appendix~\ref{appendix:variable_selection}.

\textbf{Score Normalization.} Since the memory and quantization scores have different physical meanings and scales, normalization is required for effective combination. However, simple min-max normalization reflects only local statistics and may distort the relative degree of anomaly across the entire time series. To address this, we apply Exponential Moving Average based min-max normalization separately to each score type $\ast \in \{\mathrm{mem}, \mathrm{quant}\}$:
\begin{equation}
    \mu_{\min,\ast}^{(t)} = \gamma \cdot \mu_{\min,\ast}^{(t-1)} + (1 - \gamma) \cdot \min(S_{\ast}^{(t)}),
    \quad
    \mu_{\max,\ast}^{(t)} = \gamma \cdot \mu_{\max,\ast}^{(t-1)} + (1 - \gamma) \cdot \max(S_{\ast}^{(t)}),
\end{equation}
where $\gamma$ denotes the momentum and $S_{\ast}^{(t)} \in \mathbb{R}^L$ is the per-position score of the $t$-th test batch after variable selection. Normalized score $\tilde{S}_{\ast}^{(t)} = (S_{\ast}^{(t)} - \mu_{\min,\ast}^{(t)}) / (\mu_{\max,\ast}^{(t)} - \mu_{\min,\ast}^{(t)} + \epsilon)$ accumulates past statistics while gradually incorporating new information, enabling consistent comparison of anomaly scores over time.

\textbf{Anomaly Score Aggregation.} The final anomaly score is computed as a weighted combination of the two normalized scores:
\begin{equation}
S^{(t)} 
= 
(1 - \lambda) \cdot \tilde{S}_{\text{mem}}^{(t)} 
+ 
\lambda \cdot \tilde{S}_{\text{quant}}^{(t)},
\end{equation}
where $\lambda$ is a weighting hyperparameter. The memory score measures the local scaling distance between the quantized embedding and its nearest normal prototypes in the memory bank, reflecting how far a sample deviates from normal representations. The quantization score measures the distance between the encoder output and its assigned prototype, reflecting how well a sample can be compressed into the normal prototypes. Since the two scores capture different aspects of abnormality, they provide complementary evidence for anomaly detection.

\subsection{Online Codebook Adaptation}
\label{subsec:Online-Codebook-Adaptation}

To address distribution shifts between training and inference, we propose Online Codebook Adaptation, which dynamically updates the codebook at inference time. Existing methods often filter out anomalous samples through threshold-based scoring, but such thresholds are sensitive to hyperparameter choices and susceptible to contamination from misclassified samples. In contrast, our approach leverages the fact that the codebook is trained solely on normal data to identify reliable normal samples without thresholds, and selectively updates the codebook to prevent contamination from anomalous data.

\textbf{Codebook Activation Set.} In unsupervised anomaly detection, training data are assumed to be normal. Hence, codebook entries activated during training can be regarded as prototypes of normal patterns. These entries constitute the shared memory bank $\mathcal{M}$ defined in Section~\ref{subsec:Vector-Quantized-Coreset}, and their indices at each scale are given by the activated index sets $\{\mathcal{A}_k\}_{k=1}^{K}$.

\textbf{Activation-based Pseudo-labeling.} At inference time, we generate pseudo-labels based on whether the quantization index $\hat{q}_k^{(i,j)}$ of a test sample was activated at scale $k$ during training:
\begin{equation}
    \tilde{y}_k^{(i,j)} = 
    \begin{cases} 
        0 & \text{if } \hat{q}_k^{(i,j)} \in \mathcal{A}_k \; (\text{Normal}) \\
        1 & \text{otherwise} \; (\text{Abnormal})
    \end{cases}.
\end{equation}
This enables normal sample identification without threshold tuning by treating samples assigned to training-activated entries as more reliable pseudo-normal candidates, while considering samples assigned to unused entries less reliable for adaptation.

\textbf{Contrastive Codebook Adaptation.} To encourage separation between normal and anomalous representations using pseudo-labels, we employ supervised contrastive learning. Given encoder outputs $\{\mathbf{z}_{e}^{(a)}\}$ and pseudo-labels $\{\tilde{y}^{(a)}\}$ from a test batch, where each anchor index is $a = (k,i,j)$, the contrastive loss is defined as:
\begin{equation}
    \mathcal{L}_{\text{con}} = -\sum_{a} \frac{1}{|P(a)|} \sum_{p \in P(a)} \log \frac{\exp(\mathrm{sim}(\mathbf{z}_e^{(a)}, \mathbf{z}_e^{(p)}) / \tau)}{\sum_{a' \neq a} \exp(\mathrm{sim}(\mathbf{z}_e^{(a)}, \mathbf{z}_e^{(a')}) / \tau)},
\end{equation}
where $\mathrm{sim}(\cdot, \cdot)$ denotes cosine similarity, $P(a) = \{p \mid \tilde{y}^{(p)} = \tilde{y}^{(a)}, p \neq a\}$ is the positive set sharing the same pseudo-label as anchor $a$, and $\tau$ is a temperature parameter. The overall TTA objective is:
\begin{equation}
\mathcal{L}_{\text{TTA}} =
\frac{1}{|\mathcal{I}_{\text{norm}}|}
\sum_{a \in \mathcal{I}_{\text{norm}}}
\mathcal{L}_{\text{total}}^{(a)}
+ \eta \mathcal{L}_{\text{con}},
\end{equation}
where $\mathcal{I}_{\text{norm}} = \{a \mid \tilde{y}^{(a)} = 0\}$ denotes the set of normal samples and $\eta$ is the contrastive loss weight. Notably, $\mathcal{L}_{\text{total}}$ is applied only to normal samples to prevent codebook contamination.

\textbf{Index-wise Coreset Update.}
Once the codebook is updated, each per-scale coreset is directly updated using its stored quantization indices:
\begin{equation}
    \mathcal{M}'_k = \left\{ \mathbf{c}_{k,m}^{\text{updated}} \mid m \in \mathcal{A}_k \right\},
    \quad k=1,\ldots,K,
\end{equation}
where $\mathcal{A}_k$ is the set of codebook indices activated at scale $k$ during training (cf.\ Eq.~\ref{eq:M_k}).

To prevent test data leakage, we adopt an inference-then-train strategy during TTA. For each test batch, anomaly scores are first computed with the current model state, after which the model is adapted using the same batch. This ensures that the adapted model affects only subsequent batches, enabling fair evaluation.

\section{Experiments}
\label{sec:Experiments}

\subsection{Experiment setup}
\label{subsec:Experiment-Setups}

\textbf{Datasets.}
We conduct experiments on five benchmark datasets widely used in time-series anomaly detection, covering various domains: PSM~\cite{PSM}, SWaT~\cite{SWAT}, SMAP~\cite{SMAP_MSL}, MSL~\cite{SMAP_MSL}, and WADI~\cite{WADI}. Dataset statistics are summarized in Table~\ref{appendix:dataset-statistics}, with detailed descriptions provided in Appendix~\ref{appendix:datasets}.

\textbf{Baselines.}
To validate the effectiveness of the proposed method, we compare against seven baseline models: LSTM-AE~\cite{LSTMAE}, LSTM-VAE~\cite{LSTMVAE}, USAD~\cite{USAD}, Anomaly Transformer~\cite{AnomalyTransformer}, VTT~\cite{VTT}, CATCH~\cite{CATCH}, and D3R~\cite{D3R}. Detailed descriptions of these baselines are provided in Appendix~\ref{appendix:baselines}.

\textbf{Metrics.}
To evaluate anomaly detection performance from multiple perspectives, we employ both point-wise and range-based metrics. For point-wise evaluation, we report F1-score, AUC-ROC, and AUC-PR. The F1-score is divided into F1(K=0) and F1(K=100) depending on whether point adjustment is applied~\cite{PointAdjustment}, which considers an anomaly segment as correctly detected if any point within the segment is detected. For range-based evaluation, we use Affiliation F1~\cite{affiliation}, R-AUC-ROC, R-AUC-PR (Range-based AUC), VUS-ROC, and VUS-PR (Volume Under the Surface)~\cite{VUS}. For metrics requiring a threshold, best-F1 threshold search is applied.

\subsection{Main results}
\label{subsec:Main-results}

\begin{table}[t]
\centering
\caption{Main experimental results on five anomaly detection benchmarks. We report point-wise metrics and range-based metrics to evaluate detection robustness. \textbf{Bold} indicates the best score, and \underline{underline} indicates the second best.}
\label{tab:main_results}
\renewcommand{\arraystretch}{0.9}
\resizebox{\textwidth}{!}{%
\begin{sc}
\begin{tabular}{lcccccccccc}
\toprule
\multirow{2}{*}{\textbf{Dataset}} & \multirow{2}{*}{\textbf{Method}} & \multicolumn{9}{c}{\textbf{Metric}} \\
\cmidrule(lr){3-11}
 & & \textbf{F1(K=0)} & \textbf{F1(K=100)} & \textbf{Aff-F1} & \textbf{AUC-ROC} & \textbf{AUC-PR} & \textbf{R-AUC-ROC} & \textbf{R-AUC-PR} & \textbf{VUS-ROC} & \textbf{VUS-PR} \\
\midrule
\multirow{9}{*}{\textbf{PSM}} & CATCH & \textbf{97.70} & 42.38 & 60.77 & 62.07 & 41.13 & 70.46 & 53.01 & 69.50 & 51.74 \\
 & D3R & 93.67 & 44.99 & 57.39 & 69.17 & 48.97 & 69.49 & 50.59 & 69.18 & 50.36 \\
 & VTT & 92.81 & 48.35 & 56.13 & 76.94 & 49.93 & 76.23 & 52.95 & 76.01 & 52.69 \\
 & AT & \underline{97.43} & 2.20 & 65.19 & 50.48 & 27.88 & 52.33 & 34.81 & 52.13 & 34.61 \\
 & LSTM-AE & 93.41 & 45.54 & 53.37 & 69.70 & 50.76 & 70.26 & 51.35 & 69.94 & 51.15 \\
 & LSTM-VAE & 93.12 & 44.72 & 59.38 & 66.14 & 47.28 & 66.16 & 48.46 & 65.95 & 48.27 \\
 & USAD & 91.78 & 43.44 & 56.04 & 64.84 & 44.18 & 65.71 & 46.43 & 65.40 & 46.21 \\
\cmidrule(lr){2-11}
 & \textbf{COMET (w/ TTA)} & 95.30 & \textbf{60.26} & \textbf{72.00} & \textbf{79.54} & \textbf{60.01} & \textbf{79.56} & \textbf{62.21} & \textbf{77.92} & \textbf{61.20} \\
 & \textbf{COMET (w/o TTA)} & 95.37 & \underline{60.13} & \underline{71.80} & \underline{79.15} & \underline{59.10} & \underline{79.18} & \underline{61.45} & \underline{77.50} & \underline{60.42} \\
\midrule
\multirow{9}{*}{\textbf{SWaT}} & CATCH & 91.35 & 10.11 & 69.12 & 23.78 & 8.66 & 27.34 & 9.95 & 27.17 & 9.86 \\
 & D3R & 87.10 & 76.65 & 71.74 & 83.16 & 72.95 & 65.27 & 47.41 & 65.35 & 47.57 \\
 & VTT & 86.13 & \textbf{77.05} & 60.11 & 81.42 & 71.08 & 58.43 & 43.62 & 58.48 & 43.74 \\
 & AT & \textbf{95.98} & 3.91 & 61.05 & 49.29 & 13.06 & 49.63 & 12.35 & 49.63 & 12.36 \\
 & LSTM-AE & 85.39 & \underline{76.73} & 72.46 & 81.95 & 72.56 & 63.95 & 46.72 & 64.01 & 46.75 \\
 & LSTM-VAE & 85.35 & 76.57 & 71.84 & 81.97 & 72.73 & 62.02 & 45.54 & 62.30 & 45.97 \\
 & USAD & 85.36 & 76.57 & 71.78 & 81.98 & 72.74 & 62.16 & 45.65 & 62.42 & 46.08 \\
\cmidrule(lr){2-11}
 & \textbf{COMET (w/ TTA)} & \underline{91.62} & 75.24 & \textbf{74.14} & \textbf{85.50} & \textbf{74.48} & \textbf{85.09} & \textbf{66.42} & \textbf{85.06} & \textbf{66.48} \\
 & \textbf{COMET (w/o TTA)} & 91.38 & 75.06 & \underline{72.93} & \underline{85.48} & \underline{74.33} & \underline{84.85} & \underline{65.46} & \underline{84.82} & \underline{65.50} \\
\midrule
\multirow{9}{*}{\textbf{SMAP}} & CATCH & 70.44 & 12.05 & 50.85 & 43.09 & 11.90 & 44.65 & 12.88 & 44.42 & 12.83 \\
 & D3R & 69.65 & 10.29 & 53.55 & 47.21 & 11.85 & 50.04 & 13.15 & 49.97 & 13.15 \\
 & VTT & 70.99 & 10.96 & 50.78 & 46.59 & 11.73 & 49.39 & 12.81 & 49.27 & 12.82 \\
 & AT & \textbf{96.42} & 2.21 & 67.54 & 50.03 & 12.89 & 50.44 & 14.16 & 50.43 & 14.20 \\
 & LSTM-AE & 69.87 & 8.56 & 53.55 & 39.10 & 10.31 & 42.22 & 11.38 & 42.15 & 11.40 \\
 & LSTM-VAE & 69.71 & 6.91 & 53.55 & 41.46 & 10.63 & 44.34 & 11.76 & 44.19 & 11.76 \\
 & USAD & 69.18 & 7.01 & 53.55 & 38.82 & 10.25 & 41.79 & 11.36 & 41.69 & 11.36 \\
\cmidrule(lr){2-11}
 & \textbf{COMET (w/ TTA)} & 82.30 & \underline{25.28} & \textbf{68.24} & \textbf{59.17} & \textbf{16.20} & \textbf{59.43} & \textbf{17.59} & \textbf{59.20} & \textbf{17.54} \\
 & \textbf{COMET (w/o TTA)} & \underline{82.46} & \textbf{25.40} & \underline{68.01} & \underline{59.06} & \underline{16.17} & \underline{58.69} & \underline{17.51} & \underline{58.52} & \underline{17.46} \\
\midrule
\multirow{9}{*}{\textbf{MSL}} & CATCH & 81.79 & 18.08 & 55.95 & 60.41 & 14.38 & 67.78 & \underline{20.78} & 67.05 & 20.48 \\
 & D3R & 87.81 & 18.37 & 62.56 & 53.88 & 14.75 & 61.91 & 19.97 & 61.42 & 19.72 \\
 & VTT & \underline{89.01} & 13.47 & 59.55 & 59.01 & 15.01 & 66.03 & 20.51 & 65.38 & 20.25 \\
 & AT & \textbf{93.84} & 2.88 & 66.49 & 50.65 & 10.62 & 52.24 & 14.87 & 52.23 & 14.99 \\
 & LSTM-AE & 88.30 & 17.62 & 62.60 & 53.61 & 14.05 & 60.18 & 18.33 & 59.77 & 18.20 \\
 & LSTM-VAE & 88.27 & 17.62 & 62.60 & 53.22 & 13.97 & 59.83 & 18.18 & 59.44 & 18.06 \\
 & USAD & 88.27 & 17.62 & 62.60 & 55.93 & 14.41 & 62.67 & 18.80 & 62.13 & 18.64 \\
\cmidrule(lr){2-11}
 & \textbf{COMET (w/ TTA)} & 87.44 & \underline{26.28} & \textbf{71.04} & \underline{65.40} & \underline{16.56} & \textbf{69.90} & \textbf{22.67} & \textbf{68.87} & \underline{22.25} \\
 & \textbf{COMET (w/o TTA)} & 87.27 & \textbf{26.44} & \underline{69.40} & \textbf{65.49} & \textbf{16.63} & \underline{69.27} & \textbf{22.67} & \underline{68.39} & \textbf{22.27} \\
\midrule
\multirow{9}{*}{\textbf{WADI}} & CATCH & 45.28 & 9.46 & 3.74 & 47.03 & 5.36 & 50.44 & 6.24 & 50.29 & 6.24 \\
 & D3R & 37.87 & 6.80 & 52.79 & 47.93 & 5.16 & 45.98 & 5.91 & 45.72 & 5.91 \\
 & VTT & 49.52 & 7.03 & 52.72 & 50.10 & 5.30 & 47.46 & 6.14 & 47.22 & 6.14 \\
 & AT & \textbf{91.10} & 2.53 & 53.81 & 50.93 & 5.77 & 51.02 & 6.69 & 51.07 & 6.72 \\
 & LSTM-AE & 33.34 & 7.06 & 52.72 & 47.19 & 4.96 & 45.47 & 5.76 & 45.24 & 5.77 \\
 & LSTM-VAE & 31.81 & 7.07 & 52.72 & 49.51 & 5.12 & 46.62 & 5.93 & 46.42 & 5.93 \\
 & USAD & 31.81 & 7.07 & 52.72 & 48.92 & 5.07 & 46.27 & 5.86 & 46.04 & 5.86 \\
\cmidrule(lr){2-11}
 & \textbf{COMET (w/ TTA)} & \underline{74.07} & \underline{18.92} & \underline{72.43} & \textbf{61.79} & \textbf{9.23} & \textbf{64.17} & \textbf{10.18} & \textbf{64.13} & \textbf{10.20} \\
 & \textbf{COMET (w/o TTA)} & 73.81 & \textbf{18.98} & \textbf{72.63} & \underline{61.19} & \underline{9.14} & \underline{63.64} & \underline{10.10} & \underline{63.57} & \underline{10.11} \\
\bottomrule
\end{tabular}
\end{sc}
}
\vskip -0.1in
\end{table}

Table~\ref{tab:main_results} summarizes the comparison across five benchmark datasets. Without test-time adaptation, COMET (w/o TTA) achieves the best performance on 39 out of 45 metrics against seven baselines, with consistent gains on range-based metrics. Although some baselines obtain high F1(K=0) scores under the Point Adjustment protocol, their performance drops on unadjusted and range-based metrics, whereas COMET maintains balanced performance and better captures continuous anomalous segments. Compared with COMET (w/o TTA), applying TTA further improves 35 out of 45 metrics, validating Online Codebook Adaptation, and its benefit tends to grow with the train--test distribution shift (Section~\ref{subsec:Analysis}). Despite the limited information available in streaming inference compared with overlapping training instances, COMET effectively adapts to distribution shifts at test time.

\subsection{Ablation study}
\label{subsec:Ablation-Study}

\begin{table}[t]
\caption{Ablation study results on COMET, averaged across all five benchmark datasets.
Detailed ablation results for each dataset are provided in Appendix~\ref{appendix:ablation}.
\textbf{Bold} indicates the best score, and \underline{underline} indicates the second best. Ties after rounding are ranked by the unrounded averages.}
\label{tab:ablation_comet}
\centering
\renewcommand{\arraystretch}{0.9}
\begin{sc}
\resizebox{\textwidth}{!}{%
\begin{tabular}{clccccccccc}
\toprule
\multirow{2}{*}{\textbf{Category}} 
& \multirow{2}{*}{\textbf{Method}} 
& \multicolumn{9}{c}{\textbf{Metric}} \\
\cmidrule(l){3-11}
& 
& \textbf{F1(K=0)} 
& \textbf{F1(K=100)} 
& \textbf{Aff-F1} 
& \textbf{AUC-ROC} 
& \textbf{AUC-PR} 
& \textbf{R-AUC-ROC} 
& \textbf{R-AUC-PR} 
& \textbf{VUS-ROC} 
& \textbf{VUS-PR} \\
\midrule
Architecture
& w/o Multi-Scale
& 76.45 & 36.31 & 69.48 & 63.17 & 30.78 & 63.62 & 28.66 & 63.25 & 28.59 \\
\midrule
\multirow{5}{*}{Scoring}
& w/o Quant Score
& 83.70 & 35.74 & 71.00 & 67.84 & 24.77 & 69.27 & 26.77 & 68.69 & 26.55 \\
& w/o Memory Score
& 81.55 & 37.96 & 69.19 & 65.62 & 34.65 & 64.85 & 32.88 & 64.48 & 32.78 \\
& w/o Local Scaling
& 83.92 & 39.31 & 71.00 & 68.85 & 34.08 & 70.64 & 34.52 & 70.17 & 34.27 \\
& w/o Variable Selection
& 83.65 & 38.54 & 70.97 & 68.16 & 34.31 & 68.50 & 34.47 & 67.94 & 34.31 \\
& w/o Normalization
& 76.70 & 31.97 & 58.62 & 58.48 & 29.95 & 58.12 & 28.10 & 57.83 & 27.92 \\
\midrule
\multirow{2}{*}{TTA}
& w/o TTA
& \uline{86.06} & \textbf{41.20} & 70.95 & 70.07 & 35.07 & 71.13 & 35.44 & 70.56 & 35.15 \\
& w/o Contrastive
& 86.04 & \uline{41.20} & \textbf{71.63} & \uline{70.20} & \textbf{35.31} & \uline{71.49} & \uline{35.71} & \uline{70.92} & \uline{35.42} \\
\midrule
\multicolumn{2}{c}{\textbf{COMET (Full)}}
& \textbf{86.15} & 41.20 & \uline{71.57} & \textbf{70.28} & \uline{35.30}
& \textbf{71.63} & \textbf{35.81} & \textbf{71.04} & \textbf{35.53} \\
\bottomrule
\end{tabular}
}
\end{sc}
\vspace{-0.1in}
\end{table}

To analyze the contribution of each component, we conduct an ablation study reported in Table~\ref{tab:ablation_comet}. Removing multi-scale patch encoding and using a single scale with patch size 6 and stride 3 leads to performance degradation, highlighting the importance of capturing temporal patterns at multiple resolutions. Within the scoring module, removing either the quantization score or the memory score degrades performance, with the quantization score being more critical for PR-based metrics and the memory score contributing more to ROC-based metrics, confirming their complementary roles. Removing variable selection also causes a consistent performance drop, indicating that filtering unstable variables improves the robustness of anomaly scoring. Replacing local scaling distance with standard Euclidean distance and removing score normalization both result in notable performance drops, demonstrating the importance of density-adaptive distance measurement and score normalization. Removing test-time adaptation leads to overall performance degradation, confirming its effectiveness in handling distribution shifts at inference. In addition, disabling contrastive learning during TTA slightly reduces several range-based metrics, suggesting that contrastive objectives contribute to more robust representation separation.

\subsection{Analysis}
\label{subsec:Analysis}

\begin{figure}[t]
  \centering
  \begin{subfigure}[b]{0.49\linewidth}
    \centering
    \includegraphics[width=\linewidth]{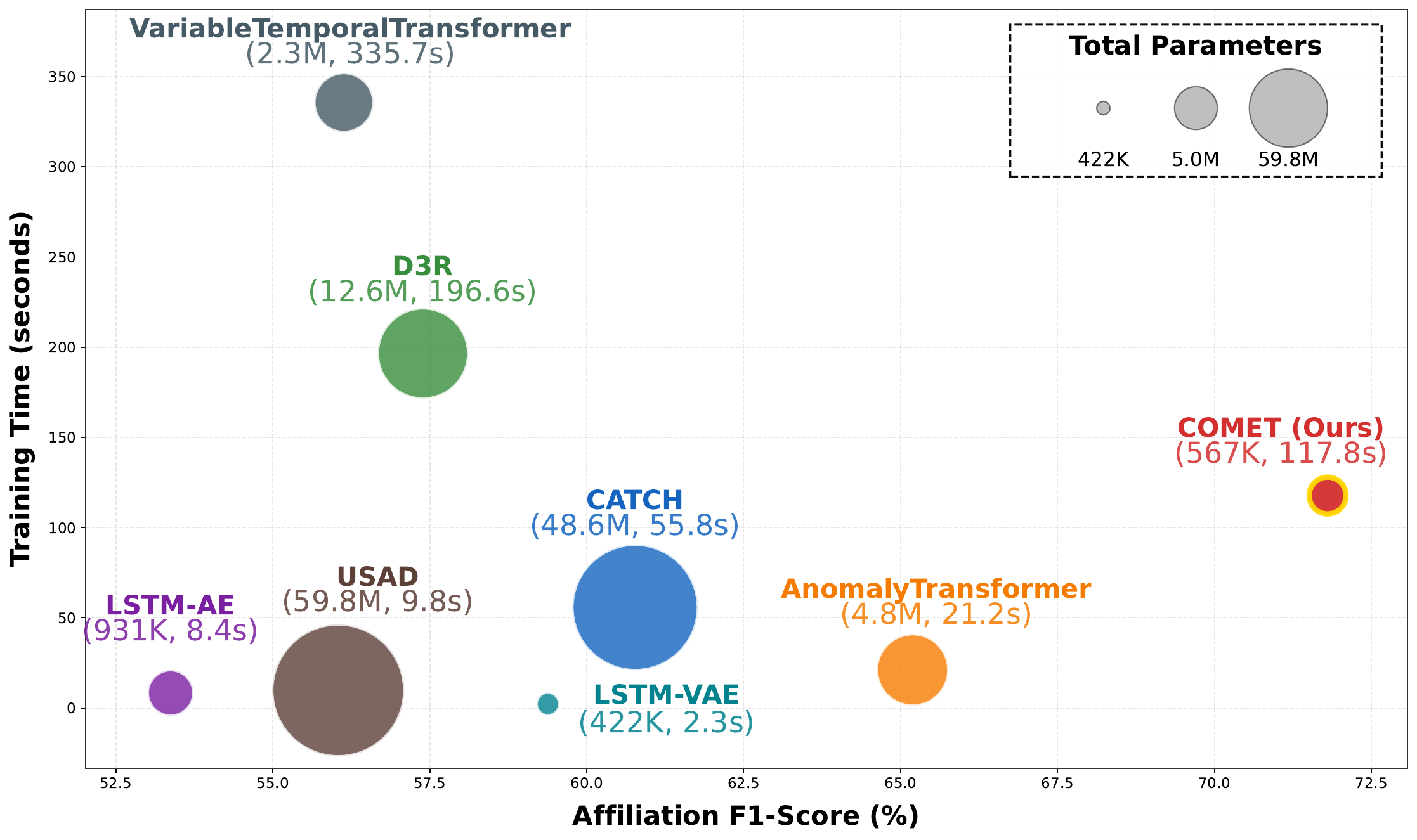}
    \caption{Parameter efficiency on PSM: COMET attains the highest Aff-F1 with a compact parameter count.}
    \label{fig:figure3}
  \end{subfigure}
  \hfill
  \begin{subfigure}[b]{0.49\linewidth}
    \centering
    \includegraphics[width=0.85\linewidth]{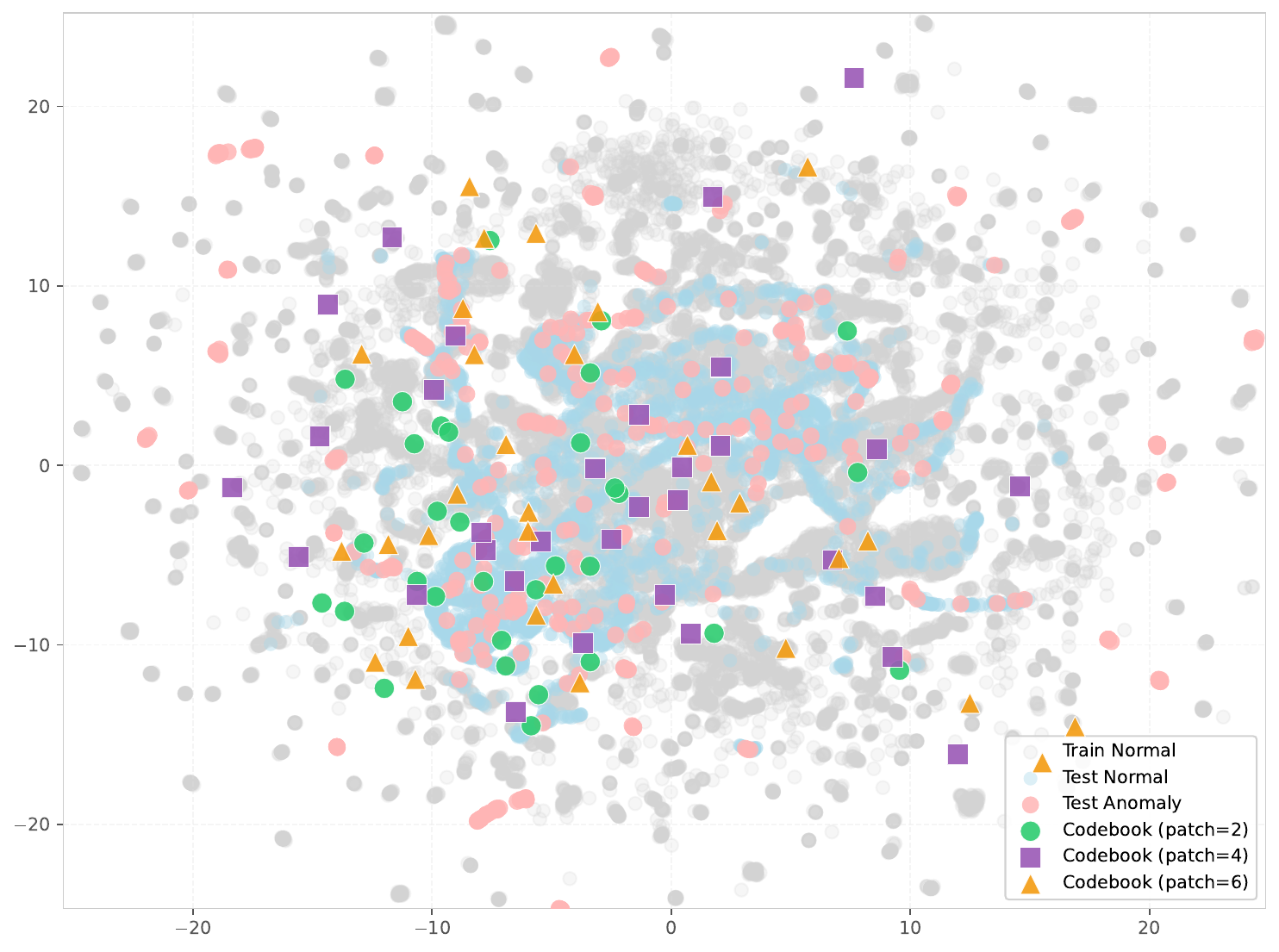}
    \caption{Embedding space on SWaT: codebook entries align with normal regions and stay separated from anomalies.}
    \label{fig:figure4}
  \end{subfigure}
  \caption{Analysis of COMET.
  \textbf{(a)} Affiliation F1-Score versus training time on the PSM dataset, where circle size indicates the number of model parameters. COMET achieves the best performance while using a compact parameter count, demonstrating its parameter efficiency relative to all baselines.
  \textbf{(b)} UMAP visualization of patch embeddings and codebook entries on the SWaT dataset. Train Normal samples are shown in gray, Test Normal in light blue, and Test Anomaly in pink. Codebook entries for each patch scale are marked as green circles (Patch 2), purple squares (Patch 4), and orange triangles (Patch 6). The codebook entries cover the normal-pattern region while remaining clearly separated from anomalies, supporting the feasibility of distance-based detection.}
  \label{fig:analysis}
  \vskip -0.18in
\end{figure}

\textbf{Efficiency Analysis.} Figure~\ref{fig:figure3} compares the Affiliation F1-score, training time, and number of parameters for each model on the PSM dataset. With only 567K parameters, COMET achieves an Affiliation F1-score of 72.00\%, outperforming larger models while using approximately 1.2\% of the parameters of CATCH (48.6M). This efficiency stems from employing lightweight linear layers and a codebook-based design rather than complex Transformer architectures. The training time of COMET is 117.8 seconds, faster than VTT (335.7 seconds) and D3R (196.6 seconds), though slower than some models due to per-scale forward passes required by multi-scale patching. Overall, COMET achieves a favorable trade-off between performance and parameter efficiency, as further illustrated by additional visualizations for all benchmark datasets in Appendix~\ref{appendix:visualizations}.

\begin{table}[t]
\caption{Learned codebook versus post-hoc memory on PSM (COMET w/o TTA). \textsc{Mem.} is the time to build the memory after training, and \textsc{Cost} is the total time cost relative to COMET.}
\label{tab:posthoc}
\centering
\renewcommand{\arraystretch}{0.9}
\begin{sc}
\resizebox{\textwidth}{!}{%
\begin{tabular}{lccccccccccc}
\toprule
\multirow{2}{*}{\textbf{Method}} & \multicolumn{9}{c}{\textbf{Metric}} & \multicolumn{2}{c}{\textbf{Cost}} \\
\cmidrule(lr){2-10}\cmidrule(lr){11-12}
& \textbf{F1(K=0)} & \textbf{F1(K=100)} & \textbf{Aff-F1} & \textbf{AUC-ROC} & \textbf{AUC-PR} & \textbf{R-AUC-ROC} & \textbf{R-AUC-PR} & \textbf{VUS-ROC} & \textbf{VUS-PR} & \textbf{Mem. (s)} & \textbf{Cost} \\
\midrule
VQEnc-KMeans & 93.28 & 50.32 & 69.44 & 71.52 & 52.92 & 69.29 & 52.83 & 68.33 & 52.19 & 677 & 4.58$\times$ \\
VQEnc-Greedy & 82.63 & 46.83 & 71.31 & 69.56 & 49.15 & 65.83 & 48.01 & 65.01 & 47.46 & \underline{25} & 1.13$\times$ \\
NoVQ-KMeans & 91.02 & 49.88 & 69.43 & 70.76 & 51.53 & 71.40 & 52.39 & 70.36 & 51.76 & 641 & 3.86$\times$ \\
NoVQ-Greedy & 83.23 & 46.65 & 71.36 & 69.49 & 48.74 & 65.81 & 47.79 & 64.98 & 47.27 & 27 & \textbf{0.61$\times$} \\
NoVQ-Contr-KMeans & \underline{94.15} & \underline{53.11} & 69.43 & \underline{73.67} & \underline{55.24} & \underline{72.42} & \underline{55.27} & \underline{70.79} & \underline{54.31} & 644 & 3.97$\times$ \\
NoVQ-Contr-Greedy & 85.37 & 49.40 & \underline{71.70} & 71.76 & 50.73 & 67.99 & 49.94 & 67.08 & 49.37 & 26 & \underline{0.70$\times$} \\
\midrule
\textbf{COMET (VQ)} & \textbf{95.37} & \textbf{60.16} & \textbf{72.00} & \textbf{79.16} & \textbf{59.11} & \textbf{79.19} & \textbf{61.46} & \textbf{77.52} & \textbf{60.43} & \textbf{6} & 1.00$\times$ \\
\bottomrule
\end{tabular}
}
\end{sc}
\vspace{-0.02in}
\end{table}

\textbf{Learned Codebook versus Post-hoc Memory.} Table~\ref{tab:posthoc} compares COMET with variants that fit an equally sized memory to all training patch embeddings after training, using KMeans or greedy $k$-center selection. VQEnc reuses the COMET encoder, NoVQ trains it without vector quantization, and NoVQ-Contr adds an InfoNCE objective. On PSM, COMET is best on all nine metrics. Post-hoc memory must store all $N$ embeddings ($O(Nd)$) and cluster them ($O(TNMd)$ for $T$ KMeans iterations), whereas COMET collects activated codewords in one pass ($O(Md)$). The KMeans variants thus take 3.9--4.6$\times$ the time of COMET, and the greedy variants lose 11--13 VUS-PR points (Appendix~\ref{appendix:posthoc}).

\textbf{Embedding Space Analysis.} Figure~\ref{fig:figure4} presents UMAP visualizations of patch embeddings and coreset entries on the SWaT dataset. The codebook entries are distributed within regions corresponding to Train Normal and Test Normal samples, indicating that representative coresets of normal patterns are effectively learned via vector quantization. Test Anomaly samples are scattered across the embedding space but reside in regions separated from the codebook entries, supporting the feasibility of anomaly detection based on memory distance and quantization error. Notably, different patch scales exhibit complementary embedding distributions, with smaller patches capturing localized patterns and larger patches spanning broader regions, confirming the benefit of multi-scale representations.

\begin{figure}[t]
  \centering
  \begin{subfigure}[b]{0.49\linewidth}
    \centering
    \includegraphics[width=0.92\linewidth]{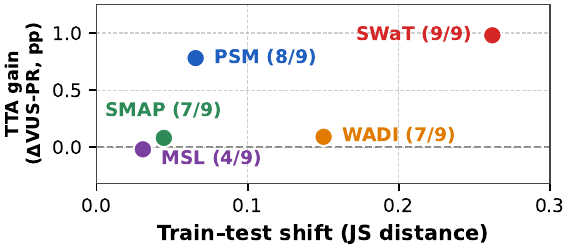}
    \caption{TTA gain vs.\ train--test shift}
    \label{fig:tta_shift}
  \end{subfigure}
  \hfill
  \begin{subfigure}[b]{0.49\linewidth}
    \centering
    \includegraphics[width=0.92\linewidth]{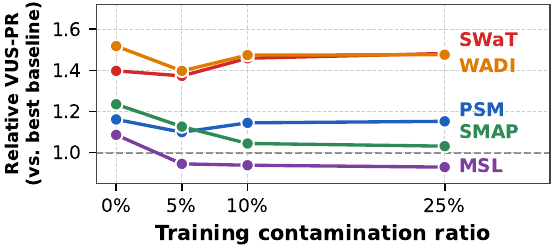}
    \caption{Robustness to training contamination}
    \label{fig:contamination}
  \end{subfigure}
  \vspace{0.02in}
  \caption{\textbf{(a)} VUS-PR gain from TTA versus train--test shift (improved metrics in parentheses). \textbf{(b)} VUS-PR of COMET (w/ TTA) relative to the best baseline. Dashed lines: no gain and parity.}
  \label{fig:shift_contam}
  \vspace{-0.25in}
\end{figure}

\textbf{TTA Gain and Distribution Shift.} We measure the shift without labels as the Jensen--Shannon distance between the training and test codeword-assignment histograms. TTA improves more metrics as the shift grows, from 4 of 9 on MSL to 9 of 9 on SWaT (Figure~\ref{fig:tta_shift}), with the largest VUS-PR gains on SWaT and PSM. On the weakly shifted MSL and SMAP, TTA is nearly neutral (Appendix~\ref{appendix:tta_shift}).

\textbf{Robustness to Training Contamination.} To measure robustness to anomaly-contaminated training data, we inject synthetic anomalies into 5--25\% of the training timesteps and retrain all models. Against the baselines, COMET keeps the best VUS-PR at every ratio on four of five datasets, by at least 3\% (Figure~\ref{fig:contamination}), while it trails the best baseline by 5--7\% on contaminated MSL (Appendix~\ref{appendix:contamination}).

\section{Conclusion}
\label{sec:Conclusion}

We propose COMET for time-series anomaly detection. COMET captures multi-scale temporal dependencies and inter-variable correlations using a Multi-scale Patch Encoder, models normal patterns as prototypes through a Vector-Quantized Coreset, and detects anomalies by combining memory distance and quantization error. To handle inference-time distribution shifts, Online Codebook Adaptation updates the codebook using codebook activation-based pseudo-labeling and contrastive learning. COMET (w/o TTA) achieves the best performance on 39 out of 45 metrics across five benchmarks compared with seven baselines, demonstrating the strength of its codebook-based representation even without test-time adaptation. However, two limitations remain: the codebook size $M$ is empirically tuned per dataset, and heavy training contamination may weaken pseudo-labeling because COMET assumes predominantly normal training data. Addressing these limitations is left as future work.

\begin{ack}
This work was supported by Institute of Information \& Communications Technology Planning \& Evaluation (IITP) grant funded by the Korea government (MSIT) (RS-2024-00460011, Climate and Environmental Data Platform for Enhancing Climate Technology Capabilities in the Anthropocene (CEDP)) and (RS-2021-II211343, Artificial Intelligence Graduate School Program (Seoul National University)). This work was also supported by the National Research Foundation of Korea (NRF) grant funded by the Korea government (MSIT) (RS-2024-00407803, RS-2025-23523657). This work was supported by the Korea Planning \& Evaluation Institute of Industrial Technology (KEIT), funded by the Ministry of Trade, Industry and Energy (MOTIE), Korea (No.~RS-2026-25509027, Development and Demonstration of an Ontology- and On-Device AI-Based Autonomous Operation Agent System for Semiconductor Wafer Manufacturing Processes). This work was also supported by the Ministry of Trade, Industry and Resources (MOTIR), Republic of Korea, in 2026, under the project ``Development of Core Foundation Model and AI Agent Technology for Thermal-Process Manufacturing (TP-MFM)'' (TPAI2616, contribution: 20\%). This work was also supported by the BK21 FOUR Program (Education and Research Center for Industrial Innovation Analytics) funded by the Ministry of Education, Korea (No.~4120240214912).
The authors declare no competing interests.
\end{ack}

\bibliographystyle{plainnat}
\bibliography{Neurips_COMET_paper}

\newpage
\appendix

\section{Method details}

\subsection{Algorithm details}
\label{appendix:algorithm}

In this appendix, we provide the detailed algorithmic procedures of COMET. Algorithm~\ref{alg:training} describes the training phase, which consists of model optimization and coreset construction. Algorithm~\ref{alg:inference} describes the inference phase with online codebook adaptation, where anomaly scores are computed before adaptation to prevent test data leakage. In the algorithms, $f_\theta^{(k)}$ denotes the Multi-scale Patch Encoder at scale $k$ (the variable-wise encoders, the core encoder, and the fusion layer), and $h_\phi^{(k)}$ denotes the decoder $h_k$ in Section~\ref{subsec:Vector-Quantized-Coreset}.

\begin{algorithm}[H]
\caption{COMET Training}
\label{alg:training}
\begin{algorithmic}
\STATE {\bfseries Input:} Training data $\mathcal{D}_{\text{train}}$, patch sizes $\mathcal{P}=\{p_1,\dots,p_K\}$, strides $\mathcal{S}=\{s_1,\dots,s_K\}$, codebook size $M$
\STATE {\bfseries Output:} Encoder $f_\theta$, decoder $h_\phi$, codebooks $\{\mathbf{C}_k\}_{k=1}^K$, per-scale coresets $\{\mathcal{M}_k\}_{k=1}^K$, shared memory bank $\mathcal{M}$, activated index sets $\{\mathcal{A}_k\}_{k=1}^K$
\vspace{0.5em}

\STATE // \textit{Phase 1: Model Training}
\FOR{each epoch}
    \FOR{each batch $\mathbf{X}\in\mathcal{D}_{\text{train}}$}
        \FOR{each scale $k=1$ {\bfseries to} $K$}
            \STATE Extract patches $\mathbf{P}_k$ with size $p_k$ and stride $s_k$
            \STATE Encode patches: $\mathbf{z}_{e,k}^{(i,j)} \leftarrow f_\theta^{(k)}(\mathbf{P}_k^{(i,j)}, \tilde{\mathbf{P}}_k^{(j)})$
            \STATE Quantize embeddings:
            \STATE \hspace{1em} $q_k^{(i,j)} \leftarrow \arg\min_{m} \|\mathbf{z}_{e,k}^{(i,j)}-\mathbf{c}_{k,m}\|_2^2$
            \STATE \hspace{1em} $\mathbf{z}_{q,k}^{(i,j)} \leftarrow \mathbf{c}_{k,q_k^{(i,j)}}$
            \STATE Decode patches: $\hat{\mathbf{P}}_k^{(i,j)} \leftarrow h_\phi^{(k)}(\mathbf{z}_{q,k}^{(i,j)})$
        \ENDFOR
        \STATE Compute $\mathcal{L}_{\text{total}}=\mathcal{L}_{\text{rec}}+\alpha\mathcal{L}_{\text{cb}}+\beta\mathcal{L}_{\text{cm}}$
        \STATE Update $\theta,\phi,\{\mathbf{C}_k\}$ via gradient descent
    \ENDFOR
\ENDFOR
\vspace{0.5em}

\STATE // \textit{Phase 2: Coreset Construction}
\STATE Initialize activated index sets $\mathcal{A}_k\leftarrow\emptyset$ for $k=1,\ldots,K$
\FOR{each batch $\mathbf{X}\in\mathcal{D}_{\text{train}}$}
    \FOR{each scale $k=1$ {\bfseries to} $K$}
        \STATE Extract patches $\mathbf{P}_k$ with size $p_k$ and stride $s_k$
        \STATE Encode patches: $\mathbf{z}_{e,k}^{(i,j)} \leftarrow f_\theta^{(k)}(\mathbf{P}_k^{(i,j)}, \tilde{\mathbf{P}}_k^{(j)})$
        \STATE Quantize embeddings:
        \STATE \hspace{1em} $q_k^{(i,j)} \leftarrow \arg\min_{m} \|\mathbf{z}_{e,k}^{(i,j)}-\mathbf{c}_{k,m}\|_2^2$
        \STATE $\mathcal{A}_k \leftarrow \mathcal{A}_k \cup \{q_k^{(i,j)}\}$
    \ENDFOR
\ENDFOR
\STATE Construct per-scale coreset $\mathcal{M}_k\leftarrow\{\mathbf{c}_{k,m}\mid m\in\mathcal{A}_k\}$ for each $k$
\STATE Construct shared memory bank $\mathcal{M}\leftarrow\bigcup_{k=1}^{K}\mathcal{M}_k$
\STATE Compute local scales $\{\sigma_{\mathbf{m}_i}\}$ for all $\mathbf{m}_i\in\mathcal{M}$
\end{algorithmic}
\end{algorithm}

The training procedure (Algorithm~\ref{alg:training}) consists of two phases. In the first phase, the model learns to encode patches into quantized representations by minimizing the combination of reconstruction loss, codebook loss, and commitment loss. In the second phase, the coreset memory bank is constructed by collecting all codebook entries activated during training, along with their local scales for density-adaptive distance computation.

\begin{algorithm}[H]
\caption{COMET Inference with Online Codebook Adaptation}
\label{alg:inference}
\begin{algorithmic}
\STATE {\bfseries Input:} Test stream $\{\mathbf{X}^{(t)}\}$, trained model $(f_\theta,h_\phi,\{\mathbf{C}_k\})$, per-scale coresets $\{\mathcal{M}_k\}_{k=1}^K$, shared memory bank $\mathcal{M}$, activated index sets $\{\mathcal{A}_k\}_{k=1}^K$
\STATE {\bfseries Output:} Anomaly scores $\{S^{(t)}\}$
\vspace{0.5em}

\FOR{each test batch $\mathbf{X}^{(t)}$}
    \STATE // \textit{Step 1: Anomaly Scoring (before adaptation)}
    \FOR{each scale $k=1$ {\bfseries to} $K$}
        \STATE Extract patches $\mathbf{P}_k$ with size $p_k$ and stride $s_k$
        \STATE Encode patches: $\mathbf{z}_{e,k}^{(i,j)} \leftarrow f_\theta^{(k)}(\mathbf{P}_k^{(i,j)}, \tilde{\mathbf{P}}_k^{(j)})$
        \STATE Quantize embeddings:
        \STATE \hspace{1em} $\hat{q}_k^{(i,j)} \leftarrow \arg\min_{m} \|\mathbf{z}_{e,k}^{(i,j)}-\mathbf{c}_{k,m}\|_2^2$
        \STATE \hspace{1em} $\mathbf{z}_{q,k}^{(i,j)} \leftarrow \mathbf{c}_{k,\,\hat{q}_k^{(i,j)}}$
        \STATE Compute query local scale $\sigma_{\mathbf{z}_{q,k}^{(i,j)}}$ using $\mathcal{N}_n(\mathbf{z}_{q,k}^{(i,j)})$ in $\mathcal{M}$
        \STATE Compute memory score:
        \STATE \hspace{1em} $S_{\text{mem},k}^{(i,j)} \leftarrow \frac{1}{n}\sum_{r=1}^{n} d_{\text{local}}(\mathbf{z}_{q,k}^{(i,j)},\mathbf{m}_r^{(k)})$, \; with $\{\mathbf{m}_r^{(k)}\}_{r=1}^{n} = \mathcal{N}_n(\mathbf{z}_{q,k}^{(i,j)})$ in $\mathcal{M}$
        \STATE Compute quantization score:
        \STATE \hspace{1em} $S_{\text{quant},k}^{(i,j)} \leftarrow \|\mathbf{z}_{e,k}^{(i,j)}-\mathbf{z}_{q,k}^{(i,j)}\|_2$
    \ENDFOR
    \STATE Aggregate per-patch scores into per-position scores $S_{\text{mem}}^{(t)}$, $S_{\text{quant}}^{(t)}$ via variable selection, patch-to-timestep mapping, and scale averaging (Section~\ref{subsec:Anomaly-Scoring} and Appendix~\ref{appendix:variable_selection})
    \STATE Apply EMA normalization
    \STATE Compute final score $S^{(t)} \leftarrow (1-\lambda)\tilde{S}_{\text{mem}}^{(t)}+\lambda\tilde{S}_{\text{quant}}^{(t)}$
\vspace{0.5em}

    \STATE // \textit{Step 2: Online Codebook Adaptation}
    \FOR{each scale $k$, variable $i$, patch index $j$}
        \IF{$\hat{q}_k^{(i,j)}\in\mathcal{A}_k$}
            \STATE $\tilde{y}_k^{(i,j)}\leftarrow 0$ \COMMENT{Normal}
        \ELSE
            \STATE $\tilde{y}_k^{(i,j)}\leftarrow 1$ \COMMENT{Abnormal}
        \ENDIF
    \ENDFOR
    \STATE $\mathcal{I}_{\text{norm}} \leftarrow \{(k,i,j) \mid \tilde{y}_k^{(i,j)}=0\}$
    \STATE Compute contrastive loss $\mathcal{L}_{\text{con}}$ over anchors $a=(k,i,j)$ using $\{\mathbf{z}_{e,k}^{(i,j)},\tilde{y}_k^{(i,j)}\}$
    \STATE $\mathcal{L}_{\text{TTA}} \leftarrow \frac{1}{|\mathcal{I}_{\text{norm}}|}\sum_{(k,i,j)\in\mathcal{I}_{\text{norm}}}\mathcal{L}_{\text{total}}^{(k,i,j)}+\eta\mathcal{L}_{\text{con}}$
    \STATE Update $\theta,\phi,\{\mathbf{C}_k\}$ via gradient descent
    \STATE Update memory bank:
    \STATE \hspace{1em} $\mathcal{M}_k\leftarrow\{\mathbf{c}_{k,m}^{\text{updated}}\mid m\in\mathcal{A}_k\}$ for each $k$
    \STATE \hspace{1em} $\mathcal{M}\leftarrow\bigcup_{k=1}^{K}\mathcal{M}_k$, and recompute local scales $\{\sigma_{\mathbf{m}_i}\}$ for all $\mathbf{m}_i\in\mathcal{M}$
    \STATE // Adaptation affects subsequent batches
\ENDFOR
\end{algorithmic}
\end{algorithm}

The inference procedure (Algorithm~\ref{alg:inference}) adopts an inference-then-train strategy to prevent test data leakage. For each test batch, anomaly scores are first computed using the current model state by combining memory distance and quantization error. Subsequently, the model is adapted using pseudo-labels generated from codebook activation patterns, where samples mapped to previously activated codebook entries are labeled as normal. The contrastive loss encourages separation between normal and abnormal representations, and the coreset is updated by directly replacing codebook entries using stored quantization indices.

\subsection{Deviation-based variable selection}
\label{appendix:variable_selection}

In this appendix, we provide detailed formulations for the deviation-based variable selection mechanism described in Section~\ref{subsec:Anomaly-Scoring}.

\subsubsection{Motivation}

Given anomaly scores $\mathbf{S} \in \mathbb{R}^{D \times N_k}$ aligned to temporal positions, where $D$ denotes the number of variables and $N_k$ the number of temporal positions corresponding to patch embeddings at scale $k$, a naive approach aggregates scores by uniformly averaging across all variables.
However, this can dilute anomaly signals due to redundant or noisy variables.
To address this issue, we propose a deviation-based variable selection strategy that selects reliable variables at each temporal position based on their standardized deviations.
Additionally, to ensure stable coverage in practice, we always include the first variable ($i=1$) in the selected set.

\subsubsection{Standardized deviation computation}

For each variable $i \in \{1, \dots, D\}$, we compute the temporal mean and standard deviation of its anomaly scores across all temporal positions of the current input window:
\begin{equation}
    \hat{\mu}^{(i)} = \frac{1}{N_k} \sum_{j=0}^{N_k-1} S^{(i,j)}, \qquad
    \varsigma^{(i)} = \sqrt{\frac{1}{N_k} \sum_{j=0}^{N_k-1} \left(S^{(i,j)} - \hat{\mu}^{(i)}\right)^2}.
\end{equation}

The standardized deviation of variable $i$ at temporal position $j$ is then computed as:
\begin{equation}
    \delta^{(i,j)} = \frac{S^{(i,j)} - \hat{\mu}^{(i)}}{\varsigma^{(i)} + \epsilon},
\end{equation}
where $\epsilon$ is a small constant for numerical stability.

\subsubsection{Position-wise variable selection}

At each temporal position $j$, we select a subset of variables $\mathcal{V}^{(j)} \subseteq \{1, \dots, D\}$ based on their absolute standardized deviation values.
Variables with smaller absolute deviations exhibit more consistent temporal behavior and are therefore considered more reliable for anomaly scoring.

\textbf{Percentile-based Selection}
Given a percentile threshold $\rho \in [0,100]$, we compute the threshold value:
\begin{equation}
    \kappa^{(j)} = Q_\rho\left( \left\{ |\delta^{(i,j)}| \right\}_{i=1}^{D} \right),
\end{equation}
where $Q_\rho(\cdot)$ denotes the $\rho$-th percentile.
The selected variable set is defined as:
\begin{equation}
    \mathcal{V}^{(j)} = \left\{ i \in \{1, \dots, D\} \mid |\delta^{(i,j)}| \le \kappa^{(j)} \right\} \cup \{1\}.
\end{equation}

\textbf{Budget-constrained Selection}
Alternatively, a fixed budget $B$ can be specified to select the $B$ most stable variables:
\begin{equation}
    \mathcal{V}^{(j)} =
    \underset{\mathcal{V} \subseteq \{1,\dots,D\},\, |\mathcal{V}|=B}{\arg\min}
    \sum_{i \in \mathcal{V}} |\delta^{(i,j)}|,
\end{equation}
where the first variable ($i=1$) is always included in $\mathcal{V}^{(j)}$.

\subsubsection{Score aggregation}

The anomaly score at temporal position $j$ is computed by averaging the scores over the selected variables:
\begin{equation}
    \bar{S}^{(j)} = \frac{1}{|\mathcal{V}^{(j)}|} \sum_{i \in \mathcal{V}^{(j)}} S^{(i,j)}.
\end{equation}
The aggregated score $\bar{S}^{(j)}$ of each patch is then mapped to timesteps, where each timestep takes the minimum score over all patches at scale $k$ that cover it, and the resulting timestep-level scores are averaged over the $K$ scales.

This selective aggregation ensures that anomaly scores are computed from variables exhibiting consistent temporal behavior, thereby improving robustness against noisy channels while preserving informative anomaly signals.

\subsubsection{Robustness against anomaly suppression}

Because Variable Selection keeps the variables with low deviation at each position, it could in principle exclude variables that respond strongly to anomalies and thereby lower recall. We examine this effect by comparing COMET with and without Variable Selection across all five benchmark datasets, reporting Precision and Recall in addition to the nine point-wise and range-based metrics used elsewhere in the paper.

\begin{table}[h]
\centering
\caption{Effect of Variable Selection across all benchmark datasets and metrics. We report Precision, Recall, and all nine point-wise and range-based metrics for both conditions, evaluated without test-time adaptation. Precision and Recall are computed under the same protocol as F1(K=0).}
\label{tab:varsel_full}
\small
\resizebox{\textwidth}{!}{%
\begin{tabular}{llccccccccccc}
\toprule
\multirow{2}{*}{\textbf{Dataset}} & \multirow{2}{*}{\textbf{Variant}} & \multicolumn{11}{c}{\textbf{Metric}} \\
\cmidrule(l){3-13}
& & \textbf{Precision} & \textbf{Recall} & \textbf{F1(K=0)} & \textbf{F1(K=100)} & \textbf{Aff-F1} & \textbf{AUC-ROC} & \textbf{AUC-PR} & \textbf{R-AUC-ROC} & \textbf{R-AUC-PR} & \textbf{VUS-ROC} & \textbf{VUS-PR} \\
\midrule
\multirow{2}{*}{PSM}  & w/o Var. Sel. & \textbf{99.10} & 89.91 & 94.28 & 59.92 & \textbf{71.85} & \textbf{79.45} & \textbf{59.44} & 73.66 & 57.74 & 72.39 & 57.04 \\
                      & COMET (w/o TTA) & 98.57 & \textbf{92.37} & \textbf{95.37} & \textbf{60.13} & 71.80 & 79.15 & 59.10 & \textbf{79.18} & \textbf{61.45} & \textbf{77.50} & \textbf{60.42} \\
\midrule
\multirow{2}{*}{SWaT} & w/o Var. Sel. & 94.85 & \textbf{93.18} & \textbf{94.01} & 74.91 & \textbf{74.94} & 85.21 & \textbf{74.51} & \textbf{85.77} & \textbf{70.52} & \textbf{85.75} & \textbf{70.67} \\
                      & COMET (w/o TTA) & \textbf{97.33} & 86.12 & 91.38 & \textbf{75.06} & 72.93 & \textbf{85.48} & 74.33 & 84.85 & 65.46 & 84.82 & 65.50 \\
\midrule
\multirow{2}{*}{SMAP} & w/o Var. Sel. & \textbf{89.83} & 64.68 & 75.20 & 21.05 & \textbf{68.09} & 57.54 & 15.34 & 56.56 & 16.63 & 56.13 & 16.58 \\
                      & COMET (w/o TTA) & 89.70 & \textbf{76.31} & \textbf{82.46} & \textbf{25.40} & 68.01 & \textbf{59.06} & \textbf{16.17} & \textbf{58.69} & \textbf{17.51} & \textbf{58.52} & \textbf{17.46} \\
\midrule
\multirow{2}{*}{MSL}  & w/o Var. Sel. & 81.90 & 72.56 & 76.95 & 19.15 & 68.05 & 57.60 & 13.35 & 63.09 & 17.62 & 62.09 & 17.46 \\
                      & COMET (w/o TTA) & \textbf{84.36} & \textbf{90.39} & \textbf{87.27} & \textbf{26.44} & \textbf{69.40} & \textbf{65.49} & \textbf{16.63} & \textbf{69.27} & \textbf{22.67} & \textbf{68.39} & \textbf{22.27} \\
\midrule
\multirow{2}{*}{WADI} & w/o Var. Sel. & \textbf{63.65} & \textbf{100.00} & \textbf{77.79} & 17.67 & 71.91 & 60.98 & 8.92 & 63.41 & 9.82 & 63.36 & 9.82 \\
                      & COMET (w/o TTA) & 58.49 & \textbf{100.00} & 73.81 & \textbf{18.98} & \textbf{72.63} & \textbf{61.19} & \textbf{9.14} & \textbf{63.64} & \textbf{10.10} & \textbf{63.57} & \textbf{10.11} \\
\bottomrule
\end{tabular}
}
\end{table}

In practice, recall increases on PSM, SMAP, and MSL and remains unchanged on WADI. Only on SWaT does recall decrease, where precision rises instead. Because $\mathcal{V}^{(j)}$ is recomputed independently at every temporal position, a variable excluded at one position is re-included as soon as its deviation returns to its typical range, so no variable is permanently discarded from scoring. Overall, Variable Selection improves F1(K=100) on all five datasets and the R-AUC and VUS metrics on all datasets except SWaT.

\subsection{Variable-wise and core encoder ablation}
\label{appendix:encoder_ablation}

To examine the contribution of each branch in the Multi-scale Patch Encoder, we evaluate two reduced variants of COMET. Core Only removes the Variable-wise Patch Encoder and uses only the shared Core Encoder, so all variables share a single encoding pathway. Variable Only removes the Core Encoder and uses only the Variable-wise Patch Encoder, so each variable is encoded independently without explicit modeling of inter-variable structure. We compare them with COMET (w/o TTA), which retains both branches.

\begin{table}[h]
\centering
\caption{Ablation of Variable-wise Encoder and Core Encoder branches across all benchmark datasets and metrics. All variants are evaluated without test-time adaptation, and \textbf{bold} indicates the best score.}
\label{tab:encoder_ablation}
\small
\resizebox{\textwidth}{!}{%
\begin{tabular}{llccccccccc}
\toprule
\multirow{2}{*}{\textbf{Dataset}} & \multirow{2}{*}{\textbf{Variant}} & \multicolumn{9}{c}{\textbf{Metric}} \\
\cmidrule(l){3-11}
& & \textbf{F1(K=0)} & \textbf{F1(K=100)} & \textbf{Aff-F1} & \textbf{AUC-ROC} & \textbf{AUC-PR} & \textbf{R-AUC-ROC} & \textbf{R-AUC-PR} & \textbf{VUS-ROC} & \textbf{VUS-PR} \\
\midrule
\multirow{3}{*}{PSM}  & Core Only & 89.21 & 46.15 & 69.49 & 66.89 & 44.62 & 68.25 & 46.52 & 67.38 & 46.12 \\
                      & Variable Only & 83.99 & 36.88 & 69.47 & 56.42 & 33.75 & 59.39 & 37.87 & 58.54 & 37.67 \\
                      & COMET (w/o TTA) & \textbf{95.37} & \textbf{60.13} & \textbf{71.80} & \textbf{79.15} & \textbf{59.10} & \textbf{79.18} & \textbf{61.45} & \textbf{77.50} & \textbf{60.42} \\
\midrule
\multirow{3}{*}{SWaT} & Core Only & 89.35 & 67.44 & 72.01 & 83.52 & 69.48 & 82.99 & 63.06 & 82.97 & 62.97 \\
                      & Variable Only & \textbf{91.95} & 56.10 & \textbf{73.46} & 83.34 & 58.50 & 84.06 & 56.48 & 83.99 & 56.39 \\
                      & COMET (w/o TTA) & 91.38 & \textbf{75.06} & 72.93 & \textbf{85.48} & \textbf{74.33} & \textbf{84.85} & \textbf{65.46} & \textbf{84.82} & \textbf{65.50} \\
\midrule
\multirow{3}{*}{SMAP} & Core Only & \textbf{91.15} & 17.64 & 66.55 & 48.17 & 12.15 & 51.06 & 13.55 & 50.93 & 13.53 \\
                      & Variable Only & 87.25 & 20.86 & \textbf{68.02} & 52.01 & 13.27 & 54.62 & 14.77 & 54.56 & 14.78 \\
                      & COMET (w/o TTA) & 82.46 & \textbf{25.40} & 68.01 & \textbf{59.06} & \textbf{16.17} & \textbf{58.69} & \textbf{17.51} & \textbf{58.52} & \textbf{17.46} \\
\midrule
\multirow{3}{*}{MSL}  & Core Only & 86.78 & 6.26 & 64.86 & 51.01 & 10.31 & 58.78 & 14.28 & 58.27 & 14.30 \\
                      & Variable Only & \textbf{92.85} & 16.89 & 68.43 & 52.09 & 11.14 & 59.63 & 15.41 & 59.24 & 15.44 \\
                      & COMET (w/o TTA) & 87.27 & \textbf{26.44} & \textbf{69.40} & \textbf{65.49} & \textbf{16.63} & \textbf{69.27} & \textbf{22.67} & \textbf{68.39} & \textbf{22.27} \\
\midrule
\multirow{3}{*}{WADI} & Core Only & 81.31 & \textbf{20.82} & 72.62 & \textbf{71.58} & \textbf{12.07} & \textbf{74.19} & \textbf{13.60} & \textbf{74.01} & \textbf{13.63} \\
                      & Variable Only & \textbf{88.95} & 15.50 & 71.33 & 63.01 & 9.38 & 66.04 & 10.66 & 65.83 & 10.66 \\
                      & COMET (w/o TTA) & 73.81 & 18.98 & \textbf{72.63} & 61.19 & 9.14 & 63.64 & 10.10 & 63.57 & 10.11 \\
\bottomrule
\end{tabular}
}
\end{table}

Removing either branch degrades most metrics on PSM, SWaT, SMAP, and MSL, although a single-branch variant attains a higher F1(K=0) on SWaT, SMAP, and MSL. The Core Only variant preserves inter-variable correlation modeling but loses the variable-specific dynamics required to capture distinct statistical patterns of individual variables. Conversely, the Variable Only variant retains variable-specific representations but lacks an explicit channel for inter-variable structure. COMET (w/o TTA), which combines both branches via the fusion layer, achieves the best 5-dataset average on eight of the nine metrics (all except F1(K=0)), confirming that variable-specific dynamics and inter-variable correlations are complementary rather than redundant. WADI is an exception, where the Core Only variant outperforms COMET (w/o TTA) on eight of the nine metrics, which is consistent with WADI's predominantly binary actuator signals offering limited variable-specific dynamics for the Variable-wise branch to exploit. Overall, these results support our two-branch encoder design.

\section{Experimental setup}

\subsection{Dataset descriptions}
\label{appendix:datasets}

We conduct experiments on five benchmark datasets that are widely used in time-series anomaly detection. The statistics of each dataset are summarized in Table~\ref{appendix:dataset-statistics}. For all datasets, the test split contains a mixture of normal and anomalous samples with ground-truth labels provided. For SWaT and WADI, the training data are collected during normal operation periods prior to attack scenarios and thus contain only normal samples, whereas the training data for SMAP, MSL, and PSM are provided without labels. For SMAP and MSL, which originally consist of multiple subdatasets, we concatenate the training and test portions from all subdatasets into a single time series to facilitate training. In our study, we reserve 10\% of the training data as a validation set.

\begin{table}[t]
  \caption{Statistics of benchmark datasets. \#TRAIN and \#VALID denote the number of samples in training and validation sets, respectively, where the validation set is split from the original training data with a ratio of 0.1. ANOMALY (\%) indicates the proportion of anomalous samples in the test set.}
  \label{appendix:dataset-statistics}
  \begin{center}
    \begin{small}
      \begin{sc}
        \begin{tabular}{lccccc}
            \toprule
                             & PSM     & SWaT    & SMAP    & MSL    & WADI       \\ 
            \midrule
            Variables        & 25      & 51      & 25      & 55     & 123        \\
            \#Train(0.9)          & 116,805 & 445,500 & 124,184 & 52,473 & 1,088,595  \\
            \#Valid(0.1)          & 12,979  & 49,500  & 13,820  & 5,844  & 120,956    \\
            \#Test & 87,841  & 449,919 & 435,826 & 73,729 & 172,801    \\
            Anomaly (\%)     & 27.76   & 12.14   & 12.84   & 10.53  & 5.71       \\
            \bottomrule
        \end{tabular}
      \end{sc}
    \end{small}
  \end{center}
  \vskip -0.1in
\end{table}

\textbf{PSM (Pooled Server Metrics)}~\cite{PSM} is a dataset collected from multiple application server nodes at eBay, consisting of 25 anonymized server metrics. PSM is released under the CC BY 4.0 license.

\textbf{SWaT (Secure Water Treatment)}~\cite{SWAT} is an industrial control system dataset collected from a water treatment testbed at the Singapore University of Technology and Design (SUTD). It contains 51 sensor and actuator variables and includes anomalies arising from cyber-attack scenarios. SWaT is released by iTrust, SUTD, for academic research purposes upon request.

\textbf{SMAP (Soil Moisture Active Passive)}~\cite{SMAP_MSL} is a telemetry dataset collected from NASA's soil moisture observation satellite. It consists of 25 variables and includes anomalous behaviors of spacecraft subsystems. Released by NASA Jet Propulsion Laboratory, SMAP is publicly available for research.

\textbf{MSL (Mars Science Laboratory)}~\cite{SMAP_MSL} is a telemetry dataset collected from NASA's Mars rover Curiosity. It consists of 55 variables and, together with SMAP, serves as a benchmark for anomaly detection in space systems released by NASA. Like SMAP, MSL is publicly available for research.

\textbf{WADI (Water Distribution)}~\cite{WADI} is a dataset collected from a water distribution testbed at SUTD and can be regarded as an extension of SWaT. It includes 123 sensor and actuator variables and consists of 14 days of normal operation followed by 2 days of attack scenarios (with 15 attacks). Like SWaT, WADI is released by iTrust, SUTD, for academic research purposes upon request.

\subsection{Baseline models}
\label{appendix:baselines}

This appendix provides brief descriptions of the baseline models used for comparison, covering reconstruction-based, probabilistic latent-variable, attention-based, frequency-domain, and diffusion-based approaches to multivariate time-series anomaly detection.

\textbf{LSTM-AE.}
LSTM-AE~\cite{LSTMAE}, also known as EncDec-AD, is one of the earliest reconstruction-based approaches for time-series anomaly detection. It employs an LSTM encoder--decoder trained exclusively on normal data to learn temporal dependencies by compressing each input sequence into a fixed-length vector representation and reconstructing the sequence from it. Anomalies are detected based on elevated reconstruction errors, under the assumption that abnormal patterns deviate from the learned normal dynamics.

\textbf{LSTM-VAE.}
LSTM-VAE~\cite{LSTMVAE} extends LSTM-AE by incorporating variational inference to model uncertainty in the latent space. Instead of learning deterministic latent embeddings, the model learns a probabilistic latent distribution, allowing it to capture variability in normal temporal patterns. Anomaly scores are derived from reconstruction errors or reconstruction likelihoods, reflecting deviations from the learned latent distribution.

\textbf{USAD.}
USAD~\cite{USAD} proposes a stable adversarial learning framework based on two autoencoders that share an encoder and are trained adversarially with asymmetric objectives. One autoencoder is trained to faithfully reconstruct normal data, while the other is optimized to amplify reconstruction discrepancies for anomalous patterns. This dual-objective design improves the separation between normal and abnormal samples compared to standard autoencoder-based reconstruction methods.

\textbf{Anomaly Transformer.}
Anomaly Transformer~\cite{AnomalyTransformer} is an attention-based model that explicitly models temporal dependencies using Transformer architectures. It introduces the concept of \emph{association discrepancy}, which measures the divergence between learned attention distributions and a learnable Gaussian prior association. Anomalies are identified based on attention inconsistency rather than reconstruction error alone, enabling the detection of subtle temporal irregularities.

\textbf{VTT.}
Variable Temporal Transformer (VTT)~\cite{VTT} focuses on modeling heterogeneous temporal dynamics and inter-variable dependencies in multivariate time series. It employs Transformer-based self-attention mechanisms to capture both temporal dependencies and correlations across variables. By learning variable-aware temporal representations, VTT enables anomaly detection based on deviations from the learned multivariate temporal structure.

\textbf{CATCH.}
CATCH~\cite{CATCH} is a frequency-aware anomaly detection framework designed to capture anomalies manifested in the spectral domain. It transforms time-series data into the frequency domain and applies frequency patching along with channel-aware modeling to learn spectral and temporal characteristics jointly. Anomalies are detected by combining the reconstruction errors in the time and frequency domains.

\textbf{D3R.}
D3R~\cite{D3R} is a diffusion-based reconstruction framework designed for multivariate time-series anomaly detection under distributional instability and drift. It combines time-series decomposition with diffusion-based reconstruction to model stable normal dynamics while mitigating the effect of drift. Anomalies are detected based on the reconstruction error between the input and its reconstruction, indicating deviations from learned normal trajectories.

\subsection{Implementation details}
\label{appendix:implementation_details}

The code, configuration files, and scripts for reproducing the main experimental results are available at \url{https://github.com/JINU6497/COMET}. The settings described below match the released configuration files and scripts.

\subsubsection{Training configuration}

All experiments are conducted with a fixed random seed of 42 for reproducibility. The input sequence length $L$ is set to 100 with a stride of 50 for sliding window segmentation (non-overlapping windows at test time). Models are trained for up to 20 epochs with early stopping (patience 5) on the validation MSE, using the AdamW optimizer with a learning rate of $10^{-4}$ and weight decay of $5 \times 10^{-4}$. The batch size is set to 128 for all datasets. We use Mean Squared Error (MSE) as the reconstruction loss. During Online Codebook Adaptation, the encoder, decoder, and codebooks are updated for one epoch on each test batch using AdamW with a learning rate of $10^{-4}$, and the contrastive loss weight $\eta$ and the temperature $\tau$ are both set to 0.5.

\subsubsection{Model architecture}

For the Multi-scale Patch Encoder, we use patch sizes $\mathcal{P} = \{2, 4, 6\}$ with corresponding strides $\mathcal{S} = \{1, 2, 3\}$. The core encoder dimension $d_c$ is set to 64. The loss coefficients for codebook loss and commitment loss are set to $\alpha = 1.0$ and $\beta = 1.0$, respectively. When the last patch at a scale does not reach the end of the window, the window is circularly padded at the end so that every timestep is covered.

\subsubsection{Anomaly scoring}

For memory score computation, we use local scaling distance with $n=10$ nearest neighbors for both density estimation and score aggregation. The score ratio $\lambda$ for combining memory score and quantization score is set to 0.5. For EMA-based score normalization, the momentum $\gamma$ is set to 0.75.

\subsubsection{Dataset-specific hyperparameters}

The codebook size $M$ and model dimension $d$ are tuned per dataset to accommodate varying data characteristics. Table~\ref{tab:hyperparameters} summarizes the dataset-specific hyperparameter settings.

\begin{table}[h]
\centering
\caption{Dataset-specific hyperparameter settings.}
\label{tab:hyperparameters}
\begin{tabular}{lcc}
\toprule
Dataset & Codebook Size ($M$) & Model Dimension ($d$) \\
\midrule
PSM & 128 & 256 \\
SWaT & 256 & 256 \\
SMAP & 128 & 128 \\
MSL & 256 & 128 \\
WADI & 32 & 64 \\
\bottomrule
\end{tabular}
\end{table}

\subsubsection{Compute resources}
\label{appendix:compute_resources}

All experiments were conducted on a single NVIDIA TITAN RTX GPU (24GB memory). Table~\ref{tab:compute_time} reports the per-dataset training and inference time of COMET for both the w/o TTA and w/ TTA variants. Both variants share the same training procedure but differ only in the inference strategy: the w/o TTA variant performs standard inference, while the w/ TTA variant additionally applies online codebook adaptation during inference.

\begin{table}[h]
\centering
\caption{Per-dataset training and inference time (in seconds) of COMET on a single NVIDIA TITAN RTX (24GB) GPU. Both variants share the same training procedure, and the w/ TTA variant additionally applies online codebook adaptation during inference.}
\label{tab:compute_time}
\begin{tabular}{lcccc}
\toprule
\multirow{2}{*}{Dataset} & \multicolumn{2}{c}{COMET (w/o TTA)} & \multicolumn{2}{c}{COMET (w/ TTA)} \\
& Training (s) & Inference (s) & Training (s) & Inference (s) \\
\midrule
PSM   & 117.8   & 13.8  & 117.8   & 17.9  \\
SWaT  & 540.7   & 146.3 & 540.7   & 194.4 \\
SMAP  & 167.1   & 49.2  & 167.1   & 67.4  \\
MSL   & 195.9   & 18.9  & 195.9   & 26.9  \\
WADI  & 2170.7  & 76.0  & 2170.7  & 115.8 \\
\midrule
\textbf{Total} & \textbf{3192.1} ($\approx$53 min) & \textbf{304.1} ($\approx$5 min) & \textbf{3192.1} ($\approx$53 min) & \textbf{422.4} ($\approx$7 min) \\
\bottomrule
\end{tabular}
\end{table}

For the main experiments of COMET (single seed, five datasets), the total compute is approximately 3496.3 seconds ($\approx$58 min, $\approx$0.97 GPU-hours) for the w/o TTA variant and approximately 3614.5 seconds ($\approx$60 min, $\approx$1.00 GPU-hours) for the w/ TTA variant. Additional ablation, sensitivity, and qualitative analyses reported in the appendices required additional compute beyond this, but each individual run remains within similar orders of magnitude.

\section{Additional results}

\subsection{Learned codebook versus post-hoc memory}
\label{appendix:posthoc}

Table~\ref{tab:posthoc_full} extends Table~\ref{tab:posthoc} to all five datasets and their average. Averaged over the datasets, COMET is best on eight of nine metrics, the KMeans variants take 3.3--4.1$\times$ the total time of COMET, and the greedy variants lose 4.5--5.4 VUS-PR points, while post-hoc memory is competitive on SWaT and WADI. All models are trained for the full 20 epochs without early stopping. \textbf{VQEnc} keeps the trained COMET encoder and replaces its codebooks. \textbf{NoVQ} trains the encoder without the quantization and commitment terms, and \textbf{NoVQ-Contr} additionally optimizes an InfoNCE objective between each window and a weakly augmented copy. For each variant, the continuous patch embeddings of the whole training set are collected, and the prototypes are fitted by KMeans (50 iterations) or greedy $k$-center selection, with as many prototypes at each scale as the number of codewords COMET activates, $|\mathcal{A}_k|$. The memory bank is then built exactly as in COMET. Memory-construction time covers embedding collection, prototype fitting, and bank construction, and \textsc{Cost} adds the encoder training time to it. Because the prototypes are fitted to all training patch embeddings at once, every embedding must be held in host memory. A patch embedding is produced for each variable, patch position, and scale, so on WADI, with 123 variables and about 1.1 million training timesteps, this amounts to about 484 million embeddings of dimension 64, or 124\,GB in 32-bit floating point. COMET instead stores only its 96 activated codewords, about 25\,KB.

\begin{table}[t]
\caption{Learned codebook versus post-hoc memory on each dataset and on average (COMET w/o TTA). \textsc{Mem.} is the time to build the memory after training in seconds, and \textsc{Cost} is the total training and memory-construction time relative to COMET. \textbf{Bold} indicates the best score, and \underline{underline} indicates the second best within each block. For \textsc{Mem.} and \textsc{Cost}, lower is better.}
\label{tab:posthoc_full}
\centering
\renewcommand{\arraystretch}{0.9}
\small
\resizebox{\textwidth}{!}{%
\begin{tabular}{llccccccccccc}
\toprule
\textbf{Dataset} & \textbf{Method} & \textbf{F1(K=0)} & \textbf{F1(K=100)} & \textbf{Aff-F1} & \textbf{AUC-ROC} & \textbf{AUC-PR} & \textbf{R-AUC-ROC} & \textbf{R-AUC-PR} & \textbf{VUS-ROC} & \textbf{VUS-PR} & \textbf{Mem. (s)} & \textbf{Cost} \\
\midrule
\multirow{7}{*}{\textbf{PSM}} & VQEnc-KMeans & 93.28 & 50.32 & 69.44 & 71.52 & 52.92 & 69.29 & 52.83 & 68.33 & 52.19 & 677 & 4.6$\times$ \\
 & VQEnc-Greedy & 82.63 & 46.83 & 71.31 & 69.56 & 49.15 & 65.83 & 48.01 & 65.01 & 47.46 & \underline{25} & 1.1$\times$ \\
 & NoVQ-KMeans & 91.02 & 49.88 & 69.43 & 70.76 & 51.53 & 71.40 & 52.39 & 70.36 & 51.76 & 641 & 3.9$\times$ \\
 & NoVQ-Greedy & 83.23 & 46.65 & 71.36 & 69.49 & 48.74 & 65.81 & 47.79 & 64.98 & 47.27 & 27 & \textbf{0.6$\times$} \\
 & NoVQ-Contr-KMeans & \underline{94.15} & \underline{53.11} & 69.43 & \underline{73.67} & \underline{55.24} & \underline{72.42} & \underline{55.27} & \underline{70.79} & \underline{54.31} & 644 & 4.0$\times$ \\
 & NoVQ-Contr-Greedy & 85.37 & 49.40 & \underline{71.70} & 71.76 & 50.73 & 67.99 & 49.94 & 67.08 & 49.37 & 26 & \underline{0.7$\times$} \\
\cmidrule(lr){2-13}
 & \textbf{COMET (VQ)} & \textbf{95.37} & \textbf{60.16} & \textbf{72.00} & \textbf{79.16} & \textbf{59.11} & \textbf{79.19} & \textbf{61.46} & \textbf{77.52} & \textbf{60.43} & \textbf{6} & 1.0$\times$ \\
\midrule
\multirow{7}{*}{\textbf{SWaT}} & VQEnc-KMeans & 91.63 & \underline{75.37} & 73.42 & 84.56 & 74.43 & 80.72 & 63.00 & 80.95 & 63.55 & 7,966 & 5.9$\times$ \\
 & VQEnc-Greedy & 91.76 & \textbf{76.02} & 75.14 & \textbf{88.23} & \textbf{77.47} & \textbf{86.94} & \underline{67.80} & \textbf{86.96} & \underline{68.10} & 2,098 & 2.3$\times$ \\
 & NoVQ-KMeans & 90.16 & 75.09 & \underline{75.28} & 85.66 & 74.80 & 85.36 & 67.27 & 85.34 & 67.51 & 7,237 & 4.9$\times$ \\
 & NoVQ-Greedy & \underline{91.90} & 75.06 & \textbf{75.61} & \underline{86.84} & 76.29 & 82.86 & 63.14 & 83.09 & 63.84 & 1,764 & \underline{1.6$\times$} \\
 & NoVQ-Contr-KMeans & 91.00 & 75.07 & 74.97 & 86.40 & 75.14 & \underline{86.29} & \textbf{67.89} & \underline{86.26} & \textbf{68.17} & 7,221 & 5.0$\times$ \\
 & NoVQ-Contr-Greedy & \textbf{92.12} & 75.14 & 75.14 & 86.78 & \underline{76.41} & 83.00 & 63.88 & 83.16 & 64.53 & \underline{1,755} & 1.7$\times$ \\
\cmidrule(lr){2-13}
 & \textbf{COMET (VQ)} & 91.32 & 75.21 & 72.71 & 85.62 & 74.66 & 85.02 & 66.13 & 84.96 & 66.08 & \textbf{65} & \textbf{1.0$\times$} \\
\midrule
\multirow{7}{*}{\textbf{SMAP}} & VQEnc-KMeans & 74.59 & 14.25 & 67.66 & 40.30 & 11.32 & 42.74 & 12.46 & 42.63 & 12.43 & 508 & 3.7$\times$ \\
 & VQEnc-Greedy & 69.28 & \underline{17.59} & 60.94 & \underline{45.56} & \underline{12.31} & \underline{47.09} & \underline{13.57} & \underline{46.88} & \underline{13.54} & \underline{15} & 1.1$\times$ \\
 & NoVQ-KMeans & 75.29 & 14.20 & \underline{67.91} & 39.98 & 11.22 & 42.34 & 12.41 & 42.21 & 12.38 & 480 & 2.9$\times$ \\
 & NoVQ-Greedy & 69.51 & 17.30 & 61.53 & 44.87 & 12.17 & 46.48 & 13.43 & 46.27 & 13.41 & 17 & \textbf{0.4$\times$} \\
 & NoVQ-Contr-KMeans & \underline{75.90} & 14.34 & 67.89 & 40.78 & 11.49 & 42.80 & 12.63 & 42.69 & 12.60 & 475 & 2.9$\times$ \\
 & NoVQ-Contr-Greedy & 69.05 & 17.49 & 61.24 & 45.08 & 12.24 & 46.61 & 13.49 & 46.39 & 13.47 & \underline{15} & \underline{0.5$\times$} \\
\cmidrule(lr){2-13}
 & \textbf{COMET (VQ)} & \textbf{82.46} & \textbf{25.40} & \textbf{68.01} & \textbf{59.07} & \textbf{16.17} & \textbf{58.69} & \textbf{17.51} & \textbf{58.52} & \textbf{17.46} & \textbf{4} & 1.0$\times$ \\
\midrule
\multirow{7}{*}{\textbf{MSL}} & VQEnc-KMeans & 84.36 & \underline{22.58} & \underline{71.37} & 60.11 & 14.75 & \underline{67.99} & \underline{20.58} & 67.20 & \underline{20.33} & 755 & 4.5$\times$ \\
 & VQEnc-Greedy & 83.91 & 21.94 & 61.42 & 62.72 & 14.86 & 66.22 & 19.88 & 65.31 & 19.63 & \underline{19} & 1.1$\times$ \\
 & NoVQ-KMeans & \underline{86.35} & 21.22 & \textbf{71.40} & 59.01 & 14.58 & 66.80 & 20.15 & 66.10 & 19.96 & 700 & 3.6$\times$ \\
 & NoVQ-Greedy & 83.64 & 21.84 & 61.66 & 62.67 & 14.83 & 66.19 & 19.82 & 65.27 & 19.59 & \underline{19} & \textbf{0.4$\times$} \\
 & NoVQ-Contr-KMeans & 86.19 & 22.49 & 70.57 & 60.43 & \underline{14.92} & 67.96 & 20.37 & \underline{67.31} & 20.13 & 701 & 3.7$\times$ \\
 & NoVQ-Contr-Greedy & 83.83 & 21.93 & 62.19 & \underline{62.86} & \underline{14.92} & 66.25 & 19.97 & 65.32 & 19.72 & \underline{19} & \underline{0.5$\times$} \\
\cmidrule(lr){2-13}
 & \textbf{COMET (VQ)} & \textbf{87.27} & \textbf{26.44} & 69.40 & \textbf{65.49} & \textbf{16.63} & \textbf{69.26} & \textbf{22.67} & \textbf{68.38} & \textbf{22.27} & \textbf{7} & 1.0$\times$ \\
\midrule
\multirow{7}{*}{\textbf{WADI}} & VQEnc-KMeans & \underline{70.36} & 13.45 & \underline{74.32} & 59.30 & 7.54 & 61.41 & 8.37 & 61.29 & 8.36 & 6,066 & 2.0$\times$ \\
 & VQEnc-Greedy & 55.57 & 6.20 & 71.32 & 41.86 & 4.64 & 42.55 & 5.08 & 42.48 & 5.08 & 743 & 1.1$\times$ \\
 & NoVQ-KMeans & 57.01 & 10.54 & \textbf{76.76} & 53.48 & 5.94 & 55.20 & 6.61 & 55.05 & 6.61 & 6,212 & 1.4$\times$ \\
 & NoVQ-Greedy & 48.40 & 7.20 & 73.84 & 45.00 & 4.85 & 46.04 & 5.37 & 45.96 & 5.37 & 785 & \textbf{0.5$\times$} \\
 & NoVQ-Contr-KMeans & 67.60 & \textbf{19.64} & 71.14 & \textbf{66.71} & \textbf{9.60} & \textbf{68.14} & \textbf{10.59} & \textbf{68.06} & \textbf{10.60} & 5,936 & 1.5$\times$ \\
 & NoVQ-Contr-Greedy & 64.86 & 6.82 & 70.88 & 47.65 & 5.11 & 48.47 & 5.73 & 48.35 & 5.73 & \underline{734} & \underline{0.6$\times$} \\
\cmidrule(lr){2-13}
 & \textbf{COMET (VQ)} & \textbf{74.65} & \underline{18.95} & 72.72 & \underline{61.35} & \underline{9.17} & \underline{63.82} & \underline{10.13} & \underline{63.75} & \underline{10.14} & \textbf{111} & 1.0$\times$ \\
\midrule
\multirow{7}{*}{\textbf{Average}} & VQEnc-KMeans & 82.84 & 35.19 & \underline{71.24} & 63.16 & 32.19 & 64.43 & 31.45 & 64.08 & 31.37 & 3,194 & 4.12$\times$ \\
 & VQEnc-Greedy & 76.63 & 33.72 & 68.03 & 61.59 & 31.69 & 61.73 & 30.87 & 61.33 & 30.76 & 580 & 1.34$\times$ \\
 & NoVQ-KMeans & 79.97 & 34.19 & \textbf{72.16} & 61.78 & 31.61 & 64.22 & 31.77 & 63.81 & 31.64 & 3,054 & 3.33$\times$ \\
 & NoVQ-Greedy & 75.34 & 33.61 & 68.80 & 61.77 & 31.38 & 61.48 & 29.91 & 61.11 & 29.90 & 522 & \textbf{0.72$\times$} \\
 & NoVQ-Contr-KMeans & \underline{82.97} & \underline{36.93} & 70.80 & \underline{65.60} & \underline{33.28} & \underline{67.52} & \underline{33.35} & \underline{67.02} & \underline{33.16} & 2,995 & 3.40$\times$ \\
 & NoVQ-Contr-Greedy & 79.05 & 34.16 & 68.23 & 62.83 & 31.88 & 62.46 & 30.60 & 62.06 & 30.56 & \underline{510} & \underline{0.80$\times$} \\
\cmidrule(lr){2-13}
 & \textbf{COMET (VQ)} & \textbf{86.21} & \textbf{41.23} & 70.97 & \textbf{70.14} & \textbf{35.15} & \textbf{71.20} & \textbf{35.58} & \textbf{70.63} & \textbf{35.28} & \textbf{39} & 1.00$\times$ \\
\bottomrule
\end{tabular}
}
\end{table}

\subsection{TTA gain and distribution shift}
\label{appendix:tta_shift}

For each dataset, every training patch (windows of length 100 with stride 50) and every test patch is assigned to its nearest codeword of the trained COMET model, and we compute the Jensen--Shannon distance between the resulting training and test assignment histograms at each scale. The Jensen--Shannon distance measures how much two distributions differ, so a larger value indicates a stronger shift between the training and test data. We report its average over the three patch scales (2, 4, and 6), which requires no labels. Table~\ref{tab:tta_shift_full} lists this shift together with the nine metrics of COMET with and without TTA from Table~\ref{tab:main_results}, ordered by the shift. Overall, the benefit of TTA grows with the degree of shift. When the training and test distributions are close, the codebook learned from the training data already represents most test patterns, so there is little left for adaptation to correct. Under a larger shift, however, normal test patterns are no longer well matched by the learned codewords, which inflates their anomaly scores and leads to false alarms. By updating the codewords with test patches that are likely to be normal, Online Codebook Adaptation realigns the codebook with the current data distribution and mitigates this inflation.

\begin{table}[t]
\caption{TTA effect versus train--test shift. \textsc{JS} is the Jensen--Shannon distance between training and test codeword-assignment histograms. \textbf{Bold} marks metrics improved by TTA, and the \textsc{Improved} column counts them.}
\label{tab:tta_shift_full}
\centering
\renewcommand{\arraystretch}{0.9}
\small
\resizebox{\textwidth}{!}{%
\begin{tabular}{lcl cccccccccc}
\toprule
\textbf{Dataset} & \textbf{JS} & \textbf{Variant} & \textbf{F1(K=0)} & \textbf{F1(K=100)} & \textbf{Aff-F1} & \textbf{AUC-ROC} & \textbf{AUC-PR} & \textbf{R-AUC-ROC} & \textbf{R-AUC-PR} & \textbf{VUS-ROC} & \textbf{VUS-PR} & \textbf{Improved} \\
\midrule
\multirow{2}{*}{\textbf{MSL}} & \multirow{2}{*}{0.031} & w/o TTA & 87.27 & 26.44 & 69.40 & 65.49 & 16.63 & 69.27 & 22.67 & 68.39 & 22.27 & \multirow{2}{*}{4/9} \\
 & & w/ TTA & \textbf{87.44} & 26.28 & \textbf{71.04} & 65.40 & 16.56 & \textbf{69.90} & 22.67 & \textbf{68.87} & 22.25 & \\
\midrule
\multirow{2}{*}{\textbf{SMAP}} & \multirow{2}{*}{0.045} & w/o TTA & 82.46 & 25.40 & 68.01 & 59.06 & 16.17 & 58.69 & 17.51 & 58.52 & 17.46 & \multirow{2}{*}{7/9} \\
 & & w/ TTA & 82.30 & 25.28 & \textbf{68.24} & \textbf{59.17} & \textbf{16.20} & \textbf{59.43} & \textbf{17.59} & \textbf{59.20} & \textbf{17.54} & \\
\midrule
\multirow{2}{*}{\textbf{PSM}} & \multirow{2}{*}{0.066} & w/o TTA & 95.37 & 60.13 & 71.80 & 79.15 & 59.10 & 79.18 & 61.45 & 77.50 & 60.42 & \multirow{2}{*}{8/9} \\
 & & w/ TTA & 95.30 & \textbf{60.26} & \textbf{72.00} & \textbf{79.54} & \textbf{60.01} & \textbf{79.56} & \textbf{62.21} & \textbf{77.92} & \textbf{61.20} & \\
\midrule
\multirow{2}{*}{\textbf{WADI}} & \multirow{2}{*}{0.150} & w/o TTA & 73.81 & 18.98 & 72.63 & 61.19 & 9.14 & 63.64 & 10.10 & 63.57 & 10.11 & \multirow{2}{*}{7/9} \\
 & & w/ TTA & \textbf{74.07} & 18.92 & 72.43 & \textbf{61.79} & \textbf{9.23} & \textbf{64.17} & \textbf{10.18} & \textbf{64.13} & \textbf{10.20} & \\
\midrule
\multirow{2}{*}{\textbf{SWaT}} & \multirow{2}{*}{0.262} & w/o TTA & 91.38 & 75.06 & 72.93 & 85.48 & 74.33 & 84.85 & 65.46 & 84.82 & 65.50 & \multirow{2}{*}{9/9} \\
 & & w/ TTA & \textbf{91.62} & \textbf{75.24} & \textbf{74.14} & \textbf{85.50} & \textbf{74.48} & \textbf{85.09} & \textbf{66.42} & \textbf{85.06} & \textbf{66.48} & \\
\bottomrule
\end{tabular}
}
\end{table}

\subsection{Robustness to training contamination}
\label{appendix:contamination}

To measure robustness when the training data are contaminated by anomalies, we inject synthetic anomalies into 5\%, 10\%, and 25\% of the training timesteps of each dataset and retrain COMET and all baselines on the contaminated data, keeping the test sets unchanged. The contaminated timesteps are split equally among four types: global and contextual point anomalies, and global and trend-shaped collective anomalies of length 20. Each event affects 20\% of the variables and sets their values three standard deviations away from the mean, where the mean and standard deviation are computed over the whole training series or, for contextual anomalies, within a window of radius 25. Trend-shaped anomalies instead add a ramp that grows to three standard deviations. We compare COMET with the seven baselines of Table~\ref{tab:main_results}. Table~\ref{tab:contamination_full} reports VUS-PR for every model and ratio, where the 0\% rows are the uncontaminated results of Table~\ref{tab:main_results}. Overall, COMET remains robust on four of the five datasets. Since the injected anomalies are sparse and heterogeneous, they have only a limited effect on the codewords, which are learned predominantly from normal patterns. Moreover, because the memory score averages distances over multiple nearest neighbors, a small number of contaminated entries in the memory bank cannot substantially alter the score. MSL is the exception, where CATCH shows greater robustness to contamination.

\begin{table}[t]
\caption{Robustness when the training data are contaminated by anomalies. VUS-PR is reported for each model after synthetic anomalies are injected into the given ratio of training timesteps. \textbf{Bold} indicates the best score, and \underline{underline} indicates the second best within each row.}
\label{tab:contamination_full}
\centering
\renewcommand{\arraystretch}{0.9}
\small
\resizebox{\textwidth}{!}{%
\begin{tabular}{llccccccccc}
\toprule
\textbf{Dataset} & \textbf{Ratio} & \textbf{CATCH} & \textbf{D3R} & \textbf{VTT} & \textbf{AT} & \textbf{LSTM-AE} & \textbf{LSTM-VAE} & \textbf{USAD} & \textbf{COMET (w/ TTA)} & \textbf{COMET (w/o TTA)} \\
\midrule
\multirow{4}{*}{\textbf{PSM}} & 0\% & 51.74 & 50.36 & 52.69 & 34.61 & 51.15 & 48.27 & 46.21 & \textbf{61.20} & \underline{60.42} \\
 & 5\% & 52.85 & 50.32 & 52.14 & 34.01 & 51.16 & 48.21 & 46.29 & \textbf{58.10} & \underline{57.95} \\
 & 10\% & 53.37 & 50.20 & 50.73 & 33.02 & 51.11 & 48.19 & 46.10 & \textbf{61.10} & \underline{60.70} \\
 & 25\% & 54.12 & 50.19 & 48.28 & 33.44 & 51.25 & 48.16 & 46.32 & \textbf{62.36} & \underline{61.33} \\
\midrule
\multirow{4}{*}{\textbf{SWaT}} & 0\% & 9.86 & 47.57 & 43.74 & 12.36 & 46.75 & 45.97 & 46.08 & \textbf{66.48} & \underline{65.50} \\
 & 5\% & 9.33 & 47.80 & 33.86 & 11.52 & 46.52 & 45.75 & 45.79 & \underline{65.63} & \textbf{70.64} \\
 & 10\% & 9.40 & 47.56 & 33.97 & 11.71 & 46.72 & 45.79 & 45.77 & \underline{69.38} & \textbf{69.89} \\
 & 25\% & 9.33 & 47.25 & 31.97 & 11.35 & 46.98 & 45.69 & 45.69 & \underline{70.02} & \textbf{71.75} \\
\midrule
\multirow{4}{*}{\textbf{SMAP}} & 0\% & 12.83 & 13.15 & 12.82 & 14.20 & 11.40 & 11.76 & 11.36 & \textbf{17.54} & \underline{17.46} \\
 & 5\% & 12.84 & 13.22 & 12.46 & 14.77 & 11.41 & 11.75 & 11.35 & \underline{16.64} & \textbf{16.75} \\
 & 10\% & 12.83 & 13.25 & 12.55 & 15.93 & 11.42 & 11.76 & 11.36 & \textbf{16.64} & \underline{16.58} \\
 & 25\% & 12.76 & 13.37 & 12.19 & 15.81 & 11.58 & 11.82 & 11.59 & \underline{16.31} & \textbf{16.75} \\
\midrule
\multirow{4}{*}{\textbf{MSL}} & 0\% & 20.48 & 19.72 & 20.25 & 14.99 & 18.20 & 18.06 & 18.64 & \underline{22.25} & \textbf{22.27} \\
 & 5\% & \textbf{20.59} & 19.85 & \underline{20.10} & 15.20 & 18.18 & 18.09 & 18.04 & 19.47 & 19.57 \\
 & 10\% & \textbf{20.46} & 19.78 & \underline{19.80} & 14.79 & 18.16 & 18.08 & 18.09 & 19.21 & 19.22 \\
 & 25\% & \textbf{20.79} & \underline{19.67} & 19.58 & 14.23 & 18.08 & 18.00 & 18.38 & 19.33 & 19.36 \\
\midrule
\multirow{4}{*}{\textbf{WADI}} & 0\% & 6.24 & 5.91 & 6.14 & 6.72 & 5.77 & 5.93 & 5.86 & \textbf{10.20} & \underline{10.11} \\
 & 5\% & 6.54 & 5.97 & 6.13 & 7.30 & 5.80 & 5.94 & 5.92 & \underline{10.20} & \textbf{10.97} \\
 & 10\% & 6.40 & 5.96 & 6.00 & 7.00 & 5.80 & 5.94 & 5.93 & \underline{10.32} & \textbf{10.85} \\
 & 25\% & 6.45 & 5.94 & 5.97 & 7.04 & 5.79 & 5.94 & 5.92 & \underline{10.39} & \textbf{10.52} \\
\bottomrule

\end{tabular}
}
\end{table}

\subsection{Additional visualizations}
\label{appendix:visualizations}

\begin{figure}[t]
  \centering
  \begin{subfigure}{0.49\linewidth}
    \centering
    \includegraphics[width=\linewidth]{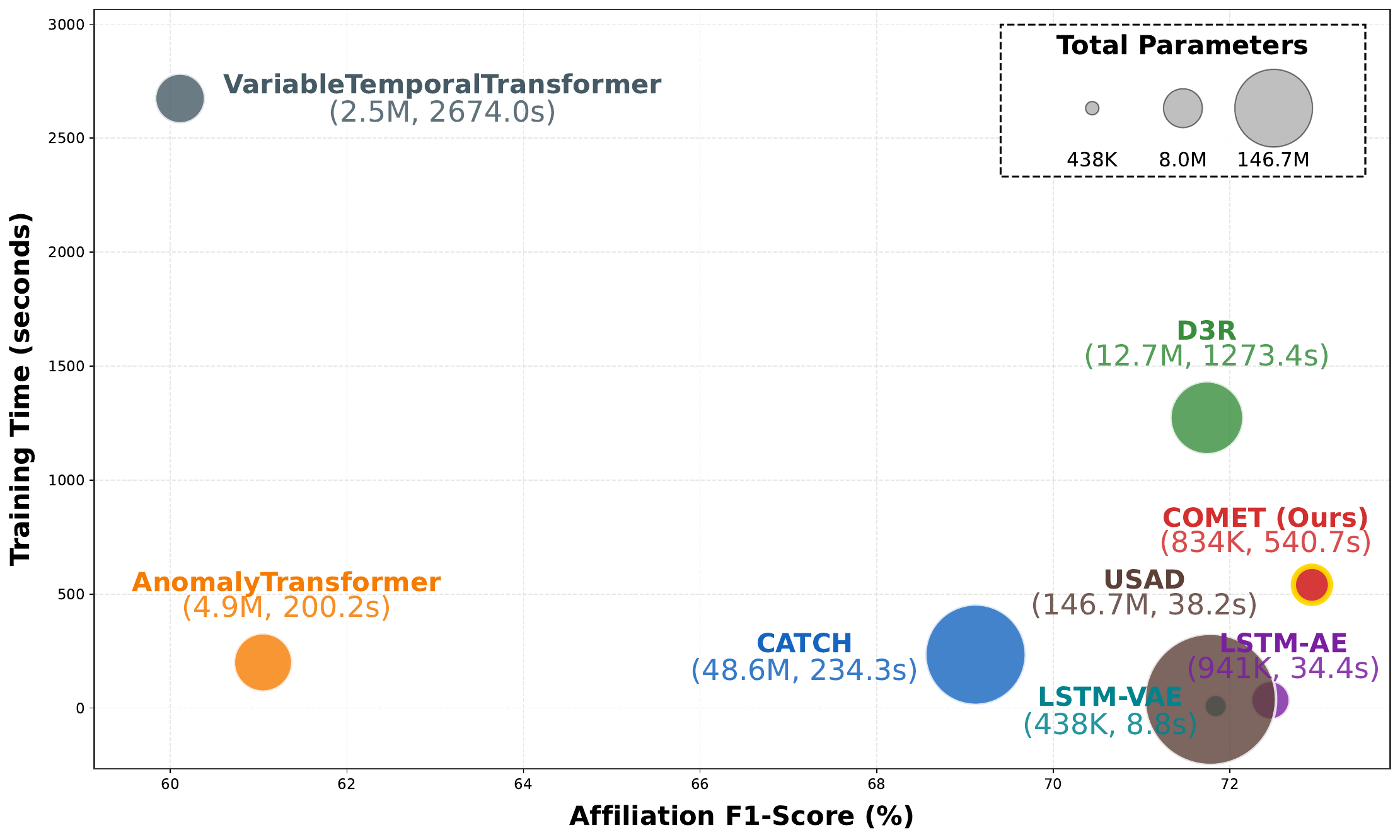}
    \caption{SWaT}
  \end{subfigure}
  \begin{subfigure}{0.49\linewidth}
    \centering
    \includegraphics[width=\linewidth]{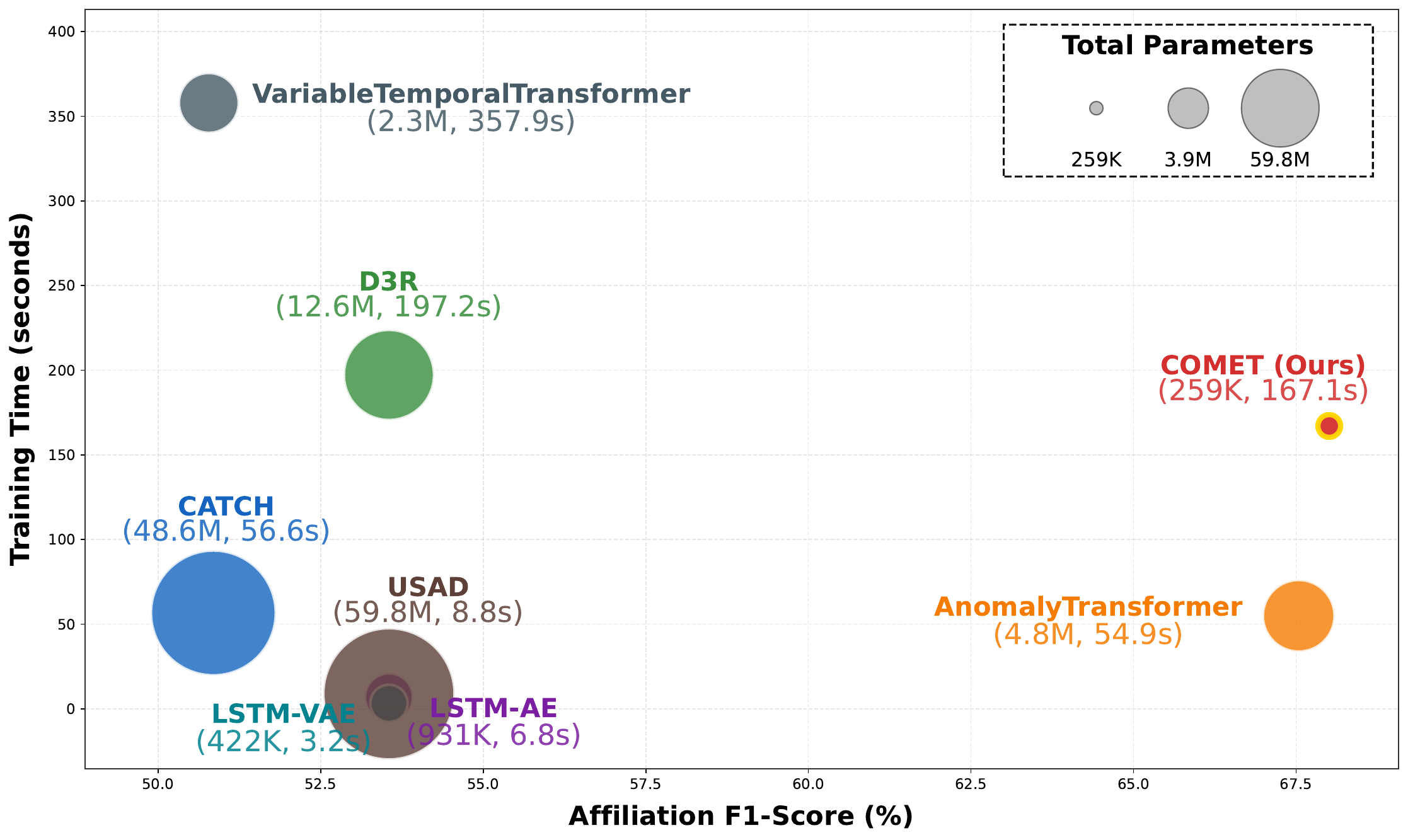}
    \caption{SMAP}
  \end{subfigure}

  \vspace{0.2em}

  \begin{subfigure}{0.49\linewidth}
    \centering
    \includegraphics[width=\linewidth]{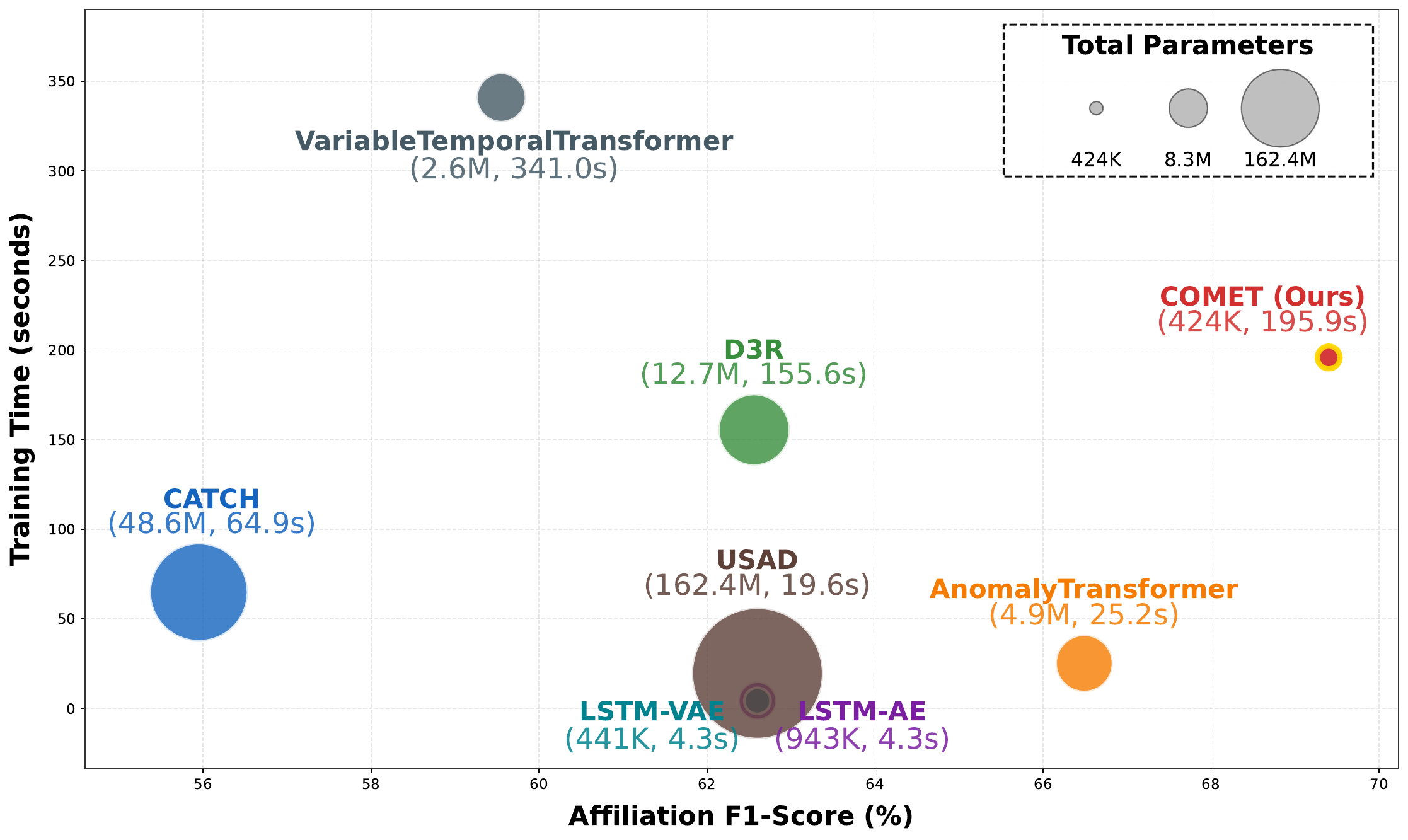}
    \caption{MSL}
  \end{subfigure}
  \begin{subfigure}{0.49\linewidth}
    \centering
    \includegraphics[width=\linewidth]{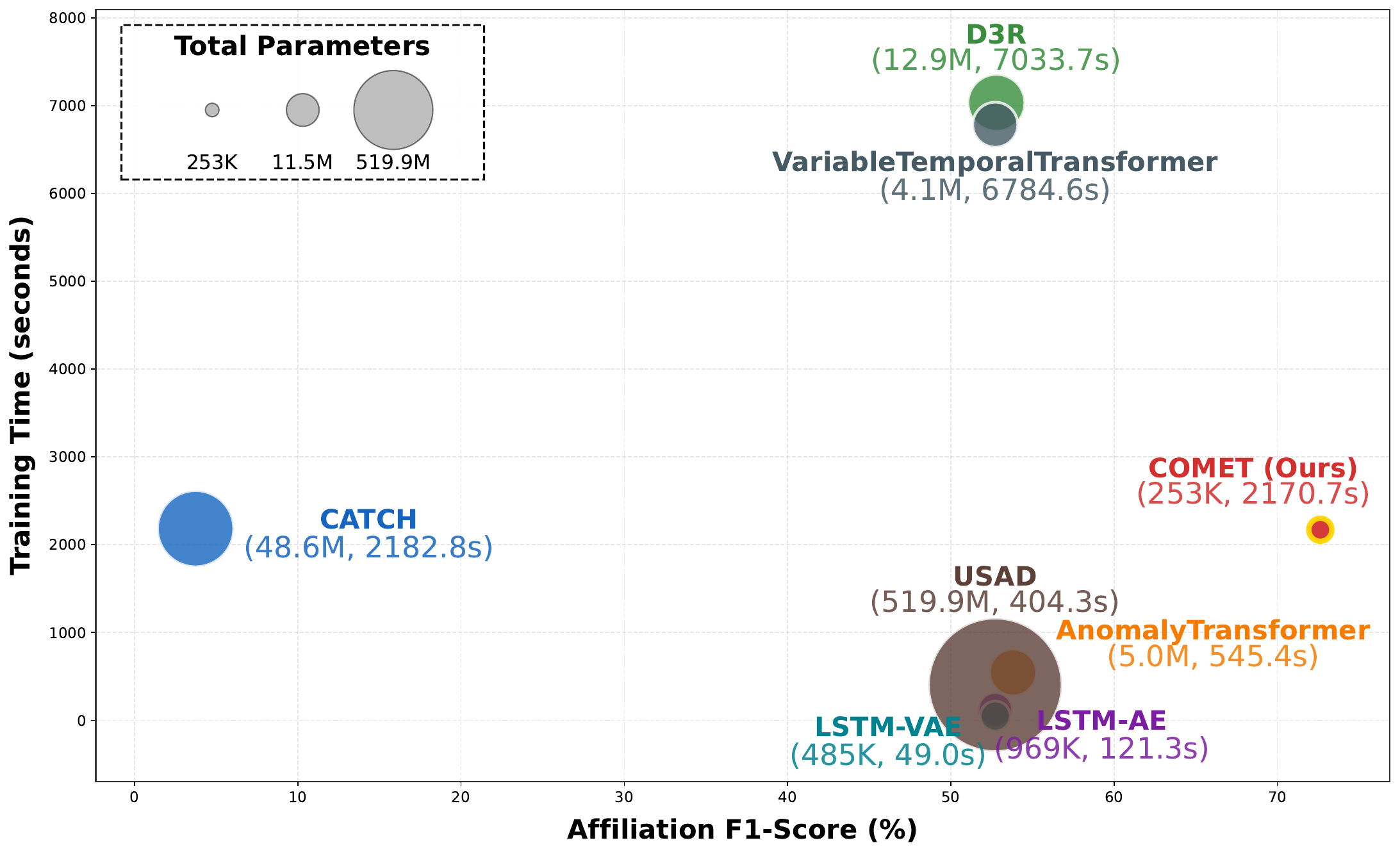}
    \caption{WADI}
  \end{subfigure}

  \caption{Performance--efficiency visualizations across datasets (Affiliation F1-score vs. training time, with circle size indicating parameter count).}
  \label{fig:appendix_visualizations}
\end{figure}

To further support the analysis of performance–efficiency trade-offs, we provide the same visualization for the remaining four benchmark datasets (PSM is shown in Figure~\ref{fig:figure3}). These figures illustrate the relationship between anomaly detection performance and training time, with marker size indicating the number of model parameters. Consistent trends across datasets demonstrate that COMET maintains strong detection performance while using a relatively small number of parameters, complementing the quantitative results reported in the main text. Figure~\ref{fig:appendix_visualizations} presents these results for the four datasets.

\subsection{Additional ablation results}
\label{appendix:ablation}

This appendix presents detailed ablation study results of COMET on each individual benchmark dataset, including PSM (Table~\ref{tab:ablation_comet_psm}), SWaT (Table~\ref{tab:ablation_comet_swat}), SMAP (Table~\ref{tab:ablation_comet_smap}), MSL (Table~\ref{tab:ablation_comet_msl}), and WADI (Table~\ref{tab:ablation_comet_wadi}). While the main paper reports results averaged across datasets for concise comparison, the per-dataset results provided here offer a more granular view of how each component contributes to performance under different data characteristics. Although the full model does not achieve the best score on every metric of every dataset, it shows strong performance on average across all datasets.

\begin{table}[t]
\caption{Ablation results of COMET on the PSM dataset. \textbf{Bold} indicates the best score, and \underline{underline} indicates the second best.}
\label{tab:ablation_comet_psm}
\centering
\renewcommand{\arraystretch}{0.9}
\small
\resizebox{\textwidth}{!}{%
\begin{tabular}{clccccccccc}
\toprule
\multirow{2}{*}{\textbf{Category}}
& \multirow{2}{*}{\textbf{Method}}
& \multicolumn{9}{c}{\textbf{Metric}} \\
\cmidrule(l){3-11}
&
& \textbf{F1(K=0)}
& \textbf{F1(K=100)}
& \textbf{Aff-F1}
& \textbf{AUC-ROC}
& \textbf{AUC-PR}
& \textbf{R-AUC-ROC}
& \textbf{R-AUC-PR}
& \textbf{VUS-ROC}
& \textbf{VUS-PR} \\
\midrule
Architecture & w/o Multi-Scale
& 87.60 & 46.49 & 71.33 & 67.04 & 48.75 & 63.50 & 46.25 & 62.71 & 45.83 \\
\midrule
\multirow{5}{*}{Scoring}
& w/o Quant Score
& \textbf{96.78} & 49.30 & 70.78 & 68.31 & 44.68 & 71.03 & 49.38 & 69.33 & 48.61 \\
& w/o Memory Score
& 94.47 & \textbf{61.68} & 73.15 & \textbf{81.40} & \textbf{64.10} & 76.66 & 61.48 & 75.58 & 60.68 \\
& w/o Local Scaling
& 95.02 & 58.11 & \textbf{73.94} & 78.46 & 57.22 & \textbf{79.76} & 60.13 & \textbf{78.24} & 59.22 \\
& w/o Variable Selection
& 94.28 & 59.92 & 71.85 & 79.45 & 59.44 & 73.66 & 57.74 & 72.39 & 57.04 \\
& w/o Normalization
& 92.92 & 45.15 & \underline{73.52} & 67.35 & 45.84 & 68.87 & 48.64 & 67.93 & 48.00 \\
\midrule
\multirow{2}{*}{TTA}
& w/o TTA
& \underline{95.37} & 60.13 & 71.80 & 79.15 & 59.10 & 79.18 & 61.45 & 77.50 & 60.42 \\
& w/o Contrastive
& 94.99 & 60.10 & 71.48 & 79.46 & \underline{60.11} & 79.39 & \underline{62.13} & 77.78 & \underline{61.14} \\
\midrule
\multicolumn{2}{c}{\textbf{COMET (Full)}}
& 95.30 & \underline{60.26} & 72.00 & \underline{79.54} & 60.01 & \underline{79.56}
& \textbf{62.21} & \underline{77.92} & \textbf{61.20} \\
\bottomrule
\end{tabular}
}
\end{table}

\begin{table}[t]
\caption{Ablation results of COMET on the SWaT dataset. \textbf{Bold} indicates the best score, and \underline{underline} indicates the second best.}
\label{tab:ablation_comet_swat}
\centering
\renewcommand{\arraystretch}{0.9}
\small
\resizebox{\textwidth}{!}{%
\begin{tabular}{clccccccccc}
\toprule
\multirow{2}{*}{\textbf{Category}}
& \multirow{2}{*}{\textbf{Method}}
& \multicolumn{9}{c}{\textbf{Metric}} \\
\cmidrule(l){3-11}
&
& \textbf{F1(K=0)}
& \textbf{F1(K=100)}
& \textbf{Aff-F1}
& \textbf{AUC-ROC}
& \textbf{AUC-PR}
& \textbf{R-AUC-ROC}
& \textbf{R-AUC-PR}
& \textbf{VUS-ROC}
& \textbf{VUS-PR} \\
\midrule
Architecture & w/o Multi-Scale
& 87.17 & 74.31 & 70.50 & 82.82 & 71.58 & 80.43 & 56.75 & 80.44 & 56.95 \\
\midrule
\multirow{5}{*}{Scoring}
& w/o Quant Score
& 70.02 & 49.52 & 71.72 & 77.32 & 31.59 & 77.17 & 29.25 & 77.07 & 29.34 \\
& w/o Memory Score
& 90.83& 75.19& \textbf{75.65} & \textbf{87.25} & \textbf{75.61} & 83.16 & 62.73 & 83.34 & 63.28 \\
& w/o Local Scaling
& \underline{91.81}& 75.14 & 73.06 & 84.24 & 73.58 & 83.79 & 65.39 & 83.75 & 65.46 \\
& w/o Variable Selection
& \textbf{94.01} & 74.91 & \underline{74.94} & 85.21& 74.51& \textbf{85.77} & \textbf{70.52} & \textbf{85.75} & \textbf{70.67} \\
& w/o Normalization
& 89.14 & \textbf{75.99} & 73.73 & 82.63 & 73.87 & 71.41 & 55.08 & 71.95 & 55.14 \\
\midrule
\multirow{2}{*}{TTA}
& w/o TTA
& 91.38 & 75.06 & 72.93 & 85.48 & 74.33 & 84.85 & 65.46 & 84.82 & 65.50 \\
& w/o Contrastive
& 91.53 & \underline{75.37}& 73.54 & \underline{85.56}& \underline{74.59}& 85.02 & 66.15 & 84.99 & 66.16 \\
\midrule
\multicolumn{2}{c}{\textbf{COMET (Full)}}
& 91.62 & 75.24 & 74.14 & 85.50 & 74.48 & \underline{85.09} & \underline{66.42} & \underline{85.06} & \underline{66.48} \\
\bottomrule
\end{tabular}
}
\end{table}

\begin{table}[t]
\caption{Ablation results of COMET on the SMAP dataset. \textbf{Bold} indicates the best score, and \underline{underline} indicates the second best.}
\label{tab:ablation_comet_smap}
\centering
\renewcommand{\arraystretch}{0.9}
\small
\resizebox{\textwidth}{!}{%
\begin{tabular}{clccccccccc}
\toprule
\multirow{2}{*}{\textbf{Category}}
& \multirow{2}{*}{\textbf{Method}}
& \multicolumn{9}{c}{\textbf{Metric}} \\
\cmidrule(l){3-11}
&
& \textbf{F1(K=0)}
& \textbf{F1(K=100)}
& \textbf{Aff-F1}
& \textbf{AUC-ROC}
& \textbf{AUC-PR}
& \textbf{R-AUC-ROC}
& \textbf{R-AUC-PR}
& \textbf{VUS-ROC}
& \textbf{VUS-PR} \\
\midrule
Architecture & w/o Multi-Scale
& 74.64 & 15.58 & 62.62 & 44.41 & 11.65 & 45.97 & 12.93 & 45.74 & 12.89 \\
\midrule
\multirow{5}{*}{Scoring}
& w/o Quant Score
& \textbf{85.63} & \textbf{26.39} & 68.17 & \textbf{60.12} & \textbf{17.01}
& \textbf{59.69} & \textbf{18.14} & \textbf{59.49} & \textbf{18.09} \\
& w/o Memory Score
& 69.55 & 18.87 & 63.97 & 45.62 & 12.70 & 46.64 & 13.88 & 46.52 & 13.85 \\
& w/o Local Scaling
& 75.16 & 24.84 & 68.09 & 58.02 & 15.59 & 59.21 & 17.10 & 59.10 & 17.04 \\
& w/o Variable Selection
& 75.20 & 21.05 & 68.09 & 57.54 & 15.34 & 56.56 & 16.63 & 56.13 & 16.58 \\
& w/o Normalization
& 69.32 & 12.82 & 57.07 & 38.87 & 10.82 & 41.21 & 11.79 & 41.09 & 11.77 \\
\midrule
\multirow{2}{*}{TTA}
& w/o TTA
& 82.46 & \underline{25.40} & 68.01 & 59.06 & 16.17
& 58.69 & 17.51 & 58.52 & 17.46 \\
& w/o Contrastive
& \underline{82.54} & 25.21 & \underline{68.18} & 58.83 & 15.98
& 59.33 & 17.38 & 59.15 & 17.32 \\
\midrule
\multicolumn{2}{c}{\textbf{COMET (Full)}}
& 82.30 & 25.28 & \textbf{68.24}
& \underline{59.17} & \underline{16.20}
& \underline{59.43} & \underline{17.59} & \underline{59.20} & \underline{17.54} \\
\bottomrule
\end{tabular}
}
\end{table}

\begin{table}[t]
\caption{Ablation results of COMET on the MSL dataset. \textbf{Bold} indicates the best score, and \underline{underline} indicates the second best.}
\label{tab:ablation_comet_msl}
\centering
\renewcommand{\arraystretch}{0.9}
\small
\resizebox{\textwidth}{!}{%
\begin{tabular}{clccccccccc}
\toprule
\multirow{2}{*}{\textbf{Category}}
& \multirow{2}{*}{\textbf{Method}}
& \multicolumn{9}{c}{\textbf{Metric}} \\
\cmidrule(l){3-11}
&
& \textbf{F1(K=0)}
& \textbf{F1(K=100)}
& \textbf{Aff-F1}
& \textbf{AUC-ROC}
& \textbf{AUC-PR}
& \textbf{R-AUC-ROC}
& \textbf{R-AUC-PR}
& \textbf{VUS-ROC}
& \textbf{VUS-PR} \\
\midrule
Architecture & w/o Multi-Scale
& 55.48 & 25.00 & 69.77 & 59.05 & 12.44 & 63.18 & 16.89 & 62.37 & 16.81 \\
\midrule
\multirow{5}{*}{Scoring}
& w/o Quant Score
& 84.42 & \textbf{33.08} & \textbf{71.76} & \textbf{69.19} & \textbf{20.20}
& \textbf{71.54} & \textbf{26.35} & \textbf{70.76} & \textbf{25.97} \\
& w/o Memory Score
& 83.35 & 24.39 & 61.24 & 63.70 & 15.01 & 66.23 & 19.93 & 65.46 & 19.71 \\
& w/o Local Scaling
& 83.21 & 20.17 & 67.96 & 63.22 & 15.28 & 67.80 & 20.34 & 67.17 & 20.01 \\
& w/o Variable Selection
& 76.95 & 19.15 & 68.05 & 57.60 & 13.35 & 63.09 & 17.62 & 62.09 & 17.46 \\
& w/o Normalization
& 82.60 & 18.95 & 35.94 & 55.93 & 14.12 & 64.13 & 19.19 & 63.41 & 18.92 \\
\midrule
\multirow{2}{*}{TTA}
& w/o TTA
& 87.27 & \underline{26.44} & 69.40 & \underline{65.49} & \underline{16.63}
& 69.27 & 22.67 & 68.39 & \underline{22.27} \\
& w/o Contrastive
& \textbf{87.48} & 26.27 & \underline{71.10} & 65.42 & 16.57
& 69.61 & 22.64 & 68.65 & 22.23 \\
\midrule
\multicolumn{2}{c}{\textbf{COMET (Full)}}
& \underline{87.44} & 26.28 & 71.04 & 65.40 & 16.56
& \underline{69.90} & \underline{22.67} & \underline{68.87} & 22.25 \\
\bottomrule
\end{tabular}
}
\end{table}

\begin{table}[t]
\caption{Ablation results of COMET on the WADI dataset. \textbf{Bold} indicates the best score, and \underline{underline} indicates the second best.}
\label{tab:ablation_comet_wadi}
\centering
\renewcommand{\arraystretch}{0.9}
\small
\resizebox{\textwidth}{!}{%
\begin{tabular}{clccccccccc}
\toprule
\multirow{2}{*}{\textbf{Category}}
& \multirow{2}{*}{\textbf{Method}}
& \multicolumn{9}{c}{\textbf{Metric}} \\
\cmidrule(l){3-11}
&
& \textbf{F1(K=0)}
& \textbf{F1(K=100)}
& \textbf{Aff-F1}
& \textbf{AUC-ROC}
& \textbf{AUC-PR}
& \textbf{R-AUC-ROC}
& \textbf{R-AUC-PR}
& \textbf{VUS-ROC}
& \textbf{VUS-PR} \\
\midrule
Architecture & w/o Multi-Scale
& 77.34 & \underline{20.19} & \underline{73.17} & \underline{62.54} & \underline{9.47}
& \underline{65.04} & \underline{10.47} & \underline{65.00} & \underline{10.48} \\
\midrule
\multirow{5}{*}{Scoring}
& w/o Quant Score
& \textbf{81.64} & \textbf{20.39} & 72.57 & \textbf{64.25} & \textbf{10.36}
& \textbf{66.91} & \textbf{10.74} & \textbf{66.81} & \textbf{10.75} \\
& w/o Memory Score
& 69.57 & 9.66 & 71.95 & 50.15 & 5.81 & 51.57 & 6.37 & 51.51 & 6.36 \\
& w/o Local Scaling
& 74.38 & 18.31 & 71.96 & 60.31 & 8.74 & 62.62 & 9.62 & 62.58 & 9.63 \\
& w/o Variable Selection
& \underline{77.79} & 17.67 & 71.91 & 60.98 & 8.92 & 63.41 & 9.82 & 63.36 & 9.82 \\
& w/o Normalization
& 49.54 & 6.96 & 52.85 & 47.62 & 5.11 & 45.00 & 5.78 & 44.76 & 5.78 \\
\midrule
\multirow{2}{*}{TTA}
& w/o TTA
& 73.81 & 18.98 & 72.63 & 61.19 & 9.14 & 63.64 & 10.10 & 63.57 & 10.11 \\
& w/o Contrastive
& 73.66 & 19.04 & \textbf{73.85} & 61.73 & 9.31 & 64.09 & 10.25 & 64.04 & 10.26 \\
\midrule
\multicolumn{2}{c}{\textbf{COMET (Full)}}
& 74.07 & 18.92 & 72.43 & 61.79 & 9.23 & 64.17 & 10.18 & 64.13 & 10.20 \\
\bottomrule
\end{tabular}
}
\end{table}

\subsection{Hyperparameter sensitivity}
\label{appendix:hyperparameter_sensitivity}

We conduct a sensitivity analysis on key hyperparameters that control the temporal granularity, representation capacity, and locality of memory-based scoring in COMET, as well as the weight that balances the two anomaly scores. Unless otherwise specified, all hyperparameters not under study are fixed to the default configuration used in the main experiments.

\subsubsection{Multi-scale patch configuration}

To analyze the effect of temporal granularity, we vary the set of patch sizes and strides used in the Multi-scale Patch Encoder. Specifically, we consider both single-scale and multi-scale configurations by combining different patch sizes and strides. The evaluated patch size sets include $\{2\}, \{4\}, \{6\}, \{2,4\}, \{4,6\}, \{2,4,6\}, \{2,4,6,8\}$, with corresponding stride configurations $\{1\}, \{2\}, \{3\}, \{1,2\}, \{2,3\}, \{1,2,3\}, \{1,2,3,4\}$. This allows us to examine how increasing temporal coverage and scale diversity affects anomaly detection performance.

As shown in Table~\ref{tab:hyper_multiscale}, combining multiple patch scales with an appropriate configuration is beneficial in most cases. The default configuration with patch sizes $\{2,4,6\}$ and strides $\{1,2,3\}$ achieves the best R-AUC and VUS metrics on PSM, SWaT, SMAP, and MSL and outperforms every single-scale setting on these datasets, suggesting that combining fine-, medium-, and coarse-grained patch scales captures complementary temporal patterns. The optimal configuration nevertheless depends on dataset characteristics: WADI, with its slower-varying actuator dynamics, benefits from a coarser single-scale setting (e.g., $\{6\}$ with stride $\{3\}$), while extending the scale set to include very large patches (e.g., $\{2,4,6,8\}$) often leads to performance saturation or degradation, suggesting that excessive temporal smoothing can dilute fine-grained anomaly signals.

\begin{table}[t]
\caption{Hyperparameter sensitivity analysis on multi-scale patch configurations (COMET w/o TTA).
Patch sizes and strides are denoted as $\{\text{patch sizes}\} \mid \{\text{strides}\}$.}
\label{tab:hyper_multiscale}
\centering
\renewcommand{\arraystretch}{0.9}
\small
\resizebox{\textwidth}{!}{%
\begin{tabular}{ccccccccccc}
\toprule
\textbf{Dataset} & \textbf{Patch \& Stride Size}
& \textbf{F1(K=0)}
& \textbf{F1(K=100)}
& \textbf{Aff-F1}
& \textbf{AUC-ROC}
& \textbf{AUC-PR}
& \textbf{R-AUC-ROC}
& \textbf{R-AUC-PR}
& \textbf{VUS-ROC}
& \textbf{VUS-PR} \\
\midrule

\multirow{7}{*}{\textbf{PSM}}
& $\{2\} \mid \{1\}$             & 93.08 & \underline{58.11} & 71.18 & \underline{77.92}& \textbf{59.17}& \underline{78.07}& \underline{59.70}& \underline{77.47}& \underline{59.20}\\
& $\{4\} \mid \{2\}$             & 92.06 & 57.90 & 70.98 & 76.78 & 53.04 & 76.30 & 56.54 & 74.22 & 55.47 \\
& $\{6\} \mid \{3\}$             & 87.60 & 46.49 & 71.33 & 67.04 & 48.75 & 63.50 & 46.25 & 62.71 & 45.83 \\
& $\{2,4\} \mid \{1,2\}$         & 82.58 & 45.25 & \underline{71.48}& 67.02 & 46.41 & 67.56 & 46.91 & 66.67 & 46.32 \\
& $\{4,6\} \mid \{2,3\}$         & 91.80 & 44.62 & 71.28 & 68.43 & 48.24 & 68.58 & 50.66 & 67.48 & 50.05 \\
& $\{2,4,6\} \mid \{1,2,3\}$     & \textbf{95.37} & \textbf{60.13} & \textbf{71.80}
                                  & \textbf{79.15} & \underline{59.10}& \textbf{79.18} & \textbf{61.45}
                                  & \textbf{77.50} & \textbf{60.42} \\
& $\{2,4,6,8\} \mid \{1,2,3,4\}$ & \underline{93.99} & 36.95 & 70.39 & 59.76 & 36.55 & 59.88 & 39.65 & 58.75 & 39.29 \\

\midrule
\multirow{7}{*}{\textbf{SWaT}}
& $\{2\} \mid \{1\}$             & \underline{89.79}& 74.15 & 71.23 & 83.39 & 72.83 & 78.64 & \underline{61.17}& 78.79 & \underline{61.33}\\
& $\{4\} \mid \{2\}$             & 85.26 & \textbf{75.85} & \underline{72.91} & \underline{84.69}& \underline{74.04}& \underline{83.13}& 60.10 & \underline{83.04}& 60.03 \\
& $\{6\} \mid \{3\}$             & 87.17 & 74.31 & 70.50 & 82.82 & 71.58 & 80.43 & 56.75 & 80.44 & 56.95 \\
& $\{2,4\} \mid \{1,2\}$         & 89.20 & \underline{75.21}& 70.08 & 82.27 & 73.15 & 72.34 & 58.39 & 72.49 & 58.19 \\
& $\{4,6\} \mid \{2,3\}$         & 87.60 & 75.20 & 70.07 & 80.84 & 71.89 & 76.49 & 56.10 & 76.58 & 56.17 \\
& $\{2,4,6\} \mid \{1,2,3\}$     & \textbf{91.38} & 75.06 & \textbf{72.93}
                                  & \textbf{85.48} & \textbf{74.33}
                                  & \textbf{84.85} & \textbf{65.46}
                                  & \textbf{84.82} & \textbf{65.50} \\
& $\{2,4,6,8\} \mid \{1,2,3,4\}$ & 87.48 & 74.65 & 71.60 & 83.82 & 73.17 & 77.02 & 55.64 & 76.97 & 55.72 \\

\midrule
\multirow{7}{*}{\textbf{SMAP}}
& $\{2\} \mid \{1\}$             & 75.99 & \underline{19.09}& 59.59 & \underline{48.23}& \underline{12.62}& 49.14 & \underline{14.00}& 49.02 & \underline{13.98}\\
& $\{4\} \mid \{2\}$             & 73.54 & 14.59 & 64.39 & 43.38 & 11.28 & 45.59 & 12.49 & 45.45 & 12.47 \\
& $\{6\} \mid \{3\}$             & 74.64 & 15.58 & 62.62 & 44.41 & 11.65 & 45.97 & 12.93 & 45.74 & 12.89 \\
& $\{2,4\} \mid \{1,2\}$         & 75.06 & 15.67 & 65.02 & 43.03 & 11.53 & 45.87 & 12.75 & 45.73 & 12.73 \\
& $\{4,6\} \mid \{2,3\}$         & \underline{78.93}& 16.21 & \underline{68.00} & 47.99 & 12.53 & \underline{50.58}& 13.78 & \underline{50.49}& 13.74 \\
& $\{2,4,6\} \mid \{1,2,3\}$     & \textbf{82.46} & \textbf{25.40} & \textbf{68.01}
                                  & \textbf{59.06} & \textbf{16.17}
                                  & \textbf{58.69} & \textbf{17.51}
                                  & \textbf{58.52} & \textbf{17.46} \\
& $\{2,4,6,8\} \mid \{1,2,3,4\}$ & 69.80 & 16.22 & 64.61 & 43.05 & 12.14 & 45.98 & 13.20 & 45.89 & 13.16 \\

\midrule
\multirow{7}{*}{\textbf{MSL}}
& $\{2\} \mid \{1\}$             & 84.81 & 17.57 & 69.05 & 57.92 & 12.62 & 64.17 & 17.16 & 63.69 & 17.06 \\
& $\{4\} \mid \{2\}$             & 81.42 & 17.67 & \textbf{71.80}& \underline{60.81}& \underline{15.07}& \underline{65.46}& \underline{19.94}& \underline{64.90}& \underline{19.68}\\
& $\{6\} \mid \{3\}$             & 55.48 & \underline{25.00}& 69.77 & 59.05 & 12.44 & 63.18 & 16.89 & 62.37 & 16.81 \\
& $\{2,4\} \mid \{1,2\}$         & \underline{86.68}& 21.95 & \underline{71.06}& 55.65 & 13.80 & 63.54 & 18.52 & 62.68 & 18.28 \\
& $\{4,6\} \mid \{2,3\}$         & 81.86 & 19.25 & 69.51 & 58.87 & 15.02 & 64.90 & 19.72 & 64.45 & 19.46 \\
& $\{2,4,6\} \mid \{1,2,3\}$     & \textbf{87.27} & \textbf{26.44} & 69.40
                                  & \textbf{65.49} & \textbf{16.63}
                                  & \textbf{69.27} & \textbf{22.67}
                                  & \textbf{68.39} & \textbf{22.27} \\
& $\{2,4,6,8\} \mid \{1,2,3,4\}$ & 78.46 & 19.65 & 69.33 & 57.23 & 14.25 & 63.83 & 18.34 & 62.98 & 18.16 \\

\midrule
\multirow{7}{*}{\textbf{WADI}}
& $\{2\} \mid \{1\}$             & \textbf{77.34}& 18.60 & 72.79 & 58.89 & 8.82 & 61.29 & 9.76 & 61.25 & 9.76 \\
& $\{4\} \mid \{2\}$             & 35.19 & 13.69 & \underline{73.08}& 55.87 & 6.23 & 57.69 & 6.90 & 57.57 & 6.90 \\
& $\{6\} \mid \{3\}$             & \textbf{77.34}& \textbf{20.19} & \textbf{73.17}
                                  & \textbf{62.54} & \textbf{9.47}
                                  & \textbf{65.04} & \textbf{10.47}
                                  & \textbf{65.00} & \textbf{10.48} \\
& $\{2,4\} \mid \{1,2\}$         & 54.91 & 11.25 & 70.82 & 51.73 & 5.93 & 54.02 & 6.60 & 53.96 & 6.60 \\
& $\{4,6\} \mid \{2,3\}$         & 50.00 & 14.71 & 71.85 & 57.45 & 6.73 & 59.47 & 7.58 & 59.40 & 7.58 \\
& $\{2,4,6\} \mid \{1,2,3\}$     & 73.81 & \underline{18.98}& 72.63 & 61.19 & \underline{9.14}& 63.64 & \underline{10.10}& 63.57 & \underline{10.11}\\
& $\{2,4,6,8\} \mid \{1,2,3,4\}$ & \underline{74.71}& 16.84 & 71.76 & \underline{61.96}& 8.75 & \underline{64.50}& 9.70 & \underline{64.49}& 9.72 \\
\bottomrule
\end{tabular}
}
\end{table}

\subsubsection{Codebook size}
We study the sensitivity to the size of the vector-quantized codebook, which controls the capacity of discrete normal pattern representations. The codebook size is varied over $\{16, 32, 64, 128, 256\}$, while keeping the multi-scale patch configuration and scoring strategy fixed. This analysis evaluates the trade-off between representational expressiveness and over-fragmentation of normal patterns.

Table~\ref{tab:hyper_codebook} shows that the best codebook size depends on the dataset. A size of 128 gives the best VUS-PR on PSM, SWaT, and SMAP, whereas MSL benefits from a larger codebook (256) and WADI from a smaller one (32). A codebook that is too small for a dataset cannot represent the diversity of its normal patterns with enough prototypes, whereas one that is too large fragments similar normal patterns into many entries and becomes more sensitive to noise. Because the appropriate capacity differs across datasets, performance is not monotonic in codebook size, and a poorly matched size can sharply lower the range-based metrics.

Although the optimal codebook size varies across datasets, this sensitivity is expected since the codebook explicitly represents prototypical normal patterns and thus reflects dataset-specific normal dynamics. Nevertheless, once an appropriate codebook size is selected for a given dataset, the model achieves strong performance on most evaluation metrics, while Aff-F1 varies little across sizes. This observation indicates that the codebook mechanism itself is robust, while its capacity should be adapted to the intrinsic complexity of the underlying data rather than fixed universally.

\begin{table}[t]
\caption{Hyperparameter sensitivity analysis on codebook size (COMET w/o TTA).}
\label{tab:hyper_codebook}
\centering
\renewcommand{\arraystretch}{0.9}
\small
\resizebox{\textwidth}{!}{%
\begin{tabular}{ccccccccccc}
\toprule
\textbf{Dataset} & \textbf{Codebook}
& \textbf{F1(K=0)}
& \textbf{F1(K=100)}
& \textbf{Aff-F1}
& \textbf{AUC-ROC}
& \textbf{AUC-PR}
& \textbf{R-AUC-ROC}
& \textbf{R-AUC-PR}
& \textbf{VUS-ROC}
& \textbf{VUS-PR} \\
\midrule

\multirow{5}{*}{\textbf{PSM}}
& 16  & 90.67 & 53.31& 71.05 & 74.38 & 50.98 & 75.70 & 53.55 & 74.87 & 53.01 \\
& 32  & 90.22 & 47.72 & 71.73 & 71.86 & 50.40 & 71.20 & 51.10 & 70.23 & 50.46 \\
& 64  & \underline{95.12} & 34.03 & 71.01 & 58.46 & 36.82 & 57.73 & 39.14 & 56.51 & 38.80 \\
& 128 & \textbf{95.37} & \textbf{60.13} & \underline{71.80}
       & \textbf{79.15} & \textbf{59.10}
       & \textbf{79.18} & \textbf{61.45}
       & \textbf{77.50} & \textbf{60.42} \\
& 256 & 94.85 & \underline{55.69}& \textbf{73.54}
       & \underline{75.43}& \underline{53.53}
       & \underline{77.20}& \underline{56.85}
       & \underline{75.83}& \underline{56.03}\\

\midrule
\multirow{5}{*}{\textbf{SWaT}}
& 16  & 91.11 & 74.64 & 70.81 & 82.01 & 72.66 & 76.05 & 59.96 & 75.83 & 59.67 \\
& 32  & 87.62 & \textbf{75.09}& 70.75 & 83.93 & 73.07 & 81.02 & 59.53 & 81.10 & 59.69 \\
& 64  & 89.86 & 74.93 & 71.10 & 82.29 & 73.01 & 73.40 & 57.12 & 73.47 & 57.10 \\
& 128 & \textbf{92.37} & 74.68 & \textbf{76.67}
       & \underline{84.06}& \underline{73.31}
       & \underline{84.21}& \textbf{67.95}
       & \underline{84.18}& \textbf{68.11} \\
& 256 & \underline{91.38} & \underline{75.06}& \underline{72.93}
       & \textbf{85.48} & \textbf{74.33}
       & \textbf{84.85} & \underline{65.46}
       & \textbf{84.82} & \underline{65.50} \\

\midrule
\multirow{5}{*}{\textbf{SMAP}}
& 16  & 78.21 & 18.42 & \textbf{68.45} & 52.75 & 13.39 & 53.03 & 14.92 & 52.78 & 14.87 \\
& 32  & \textbf{89.54} & \underline{22.08}& \underline{68.14}
       & \underline{53.14}& \underline{14.08}
       & \underline{55.94}& \underline{15.52}
       & \underline{55.71}& \underline{15.46}\\
& 64  & 78.39 & 19.47 & 67.99 & 52.24 & 14.00 & 54.74 & 15.33 & 54.62 & 15.29 \\
& 128 & \underline{82.46} & \textbf{25.40} & 68.01
       & \textbf{59.06} & \textbf{16.17}
       & \textbf{58.69} & \textbf{17.51}
       & \textbf{58.52} & \textbf{17.46} \\
& 256 & 69.88 & 20.07 & 68.03 & 50.88 & 13.57 & 52.85 & 14.85 & 52.71 & 14.81 \\

\midrule
\multirow{5}{*}{\textbf{MSL}}
& 16  & \underline{86.56} & 18.68 & \textbf{71.71} & 60.10 & 14.15 & 66.04 & 18.79 & 65.31 & 18.59 \\
& 32  & 54.36 & 16.94 & 70.74 & 51.92 & 11.83 & 58.98 & 15.32 & 58.28 & 15.18 \\
& 64  & 83.64 & 17.81 & \underline{71.25}
       & \underline{60.33}& \underline{14.25}
       & \underline{66.62}& \underline{19.31}
       & \underline{65.38}& \underline{19.03}\\
& 128 & 78.81 & \underline{20.85}& 69.59 & 55.78 & 13.19 & 62.72 & 17.74 & 62.01 & 17.55 \\
& 256 & \textbf{87.27} & \textbf{26.44} & 69.40
       & \textbf{65.49} & \textbf{16.63}
       & \textbf{69.27} & \textbf{22.67}
       & \textbf{68.39} & \textbf{22.27} \\

\midrule
\multirow{5}{*}{\textbf{WADI}}
& 16  & 63.98 & 9.74 & \underline{73.39} & 47.39 & 5.28 & 49.90 & 5.90 & 49.82 & 5.89 \\
& 32  & \textbf{73.81}& \textbf{18.98} & 72.63
       & \textbf{61.19} & \textbf{9.14}
       & \textbf{63.64} & \textbf{10.10}
       & \textbf{63.57} & \textbf{10.11} \\
& 64  & 65.04 & 7.15 & \textbf{74.51} & 39.76 & 4.61 & 41.84 & 5.09 & 41.79 & 5.09 \\
& 128 & \underline{67.69}& \underline{15.57}& 73.21
       & \underline{55.24}& \underline{6.88}
       & \underline{57.75}& \underline{7.67}
       & \underline{57.69}& \underline{7.67}\\
& 256 & 59.85 & 7.41 & 73.09 & 41.09 & 4.50 & 42.79 & 4.90 & 42.75 & 4.90 \\

\bottomrule
\end{tabular}
}
\end{table}

\subsubsection{Number of nearest neighbors}

We analyze the impact of the number of nearest neighbors used in memory-based retrieval for anomaly scoring. The number of neighbors is varied over $\{1, 3, 5, 10, 20\}$. This parameter controls the locality of neighborhood-based normalization and affects the stability of anomaly scores.

As reported in Table~\ref{tab:hyper_knn}, the best number of neighbors depends on the dataset. A very small neighborhood makes the memory score depend on only a few memory entries and is therefore sensitive to local noise, and using a single neighbor ($n=1$) yields the lowest VUS-PR on PSM, SWaT, SMAP, and WADI. A larger neighborhood smooths the score but can blur fine-grained differences between normal and anomalous patterns. Intermediate values give the best results on PSM ($n=10$), SWaT ($n=5$), and MSL ($n=5$), whereas SMAP favors a larger neighborhood ($n=20$) and WADI favors $n=3$ on the PR-based metrics and $n=20$ on the ROC-based metrics. This indicates that a moderate local neighborhood generally balances robustness and sensitivity, while the best balance depends on the noise level and local density structure of each dataset.

The default $n=10$ is the best or second best in VUS-PR on PSM, SWaT, SMAP, and MSL and close to the best on WADI, so it serves as a reasonable default without dataset-specific tuning, although an unsuitable value (e.g., $n=3$ on MSL) can noticeably lower performance.

\begin{table}[t]
\caption{Hyperparameter sensitivity analysis on the number of nearest neighbors (COMET w/o TTA).}
\label{tab:hyper_knn}
\centering
\renewcommand{\arraystretch}{0.9}
\small
\resizebox{\textwidth}{!}{%
\begin{tabular}{ccccccccccc}
\toprule
\textbf{Dataset} & \textbf{\#NN}
& \textbf{F1(K=0)}
& \textbf{F1(K=100)}
& \textbf{Aff-F1}
& \textbf{AUC-ROC}
& \textbf{AUC-PR}
& \textbf{R-AUC-ROC}
& \textbf{R-AUC-PR}
& \textbf{VUS-ROC}
& \textbf{VUS-PR} \\
\midrule

\multirow{5}{*}{\textbf{PSM}}
& 1  & \textbf{96.05} & 30.82 & 70.09 & 56.10 & 36.05 & 58.65 & 39.68 & 57.54 & 39.20 \\
& 3  & 94.38 & 35.30 & 70.10 & 57.42 & 38.90 & 60.00 & 42.04 & 58.76 & 41.44 \\
& 5  & 94.55 & \underline{56.57} & \textbf{73.79}
       & \underline{75.33}& \underline{58.70}
       & \underline{75.41}& \underline{59.97}
       & \underline{74.20}& \underline{59.07}\\
& 10 & \underline{95.37} & \textbf{60.13} & 71.80
       & \textbf{79.15} & \textbf{59.10}
       & \textbf{79.18} & \textbf{61.45}
       & \textbf{77.50} & \textbf{60.42} \\
& 20 & 93.03 & 52.07 & \underline{71.96} & 71.82 & 52.36 & 73.87 & 55.08 & 72.71 & 54.31 \\

\midrule
\multirow{5}{*}{\textbf{SWaT}}
& 1  & 89.36 & 74.71 & 72.30 & 82.84 & 72.20 & 79.74 & 60.43 & 79.71 & 60.36 \\
& 3  & 89.44 & \textbf{75.35}& 72.26 & 85.40 & 73.89 & 84.52 & 65.33 & 84.37 & 64.85 \\
& 5  & \textbf{93.96} & 74.95 & \textbf{74.29}
       & 85.09 & \underline{74.27}
       & \textbf{85.62} & \textbf{70.48}
       & \textbf{85.58} & \textbf{70.44} \\
& 10 & \underline{91.38} & \underline{75.06}& \underline{72.93}
       & \textbf{85.48} & \textbf{74.33}
       & \underline{84.85}& \underline{65.46}
       & \underline{84.82}& \underline{65.50}\\
& 20 & 89.64 & 75.05 & 72.35 & \underline{85.44}& 74.02 & 84.35 & 63.81 & 84.26 & 63.67 \\

\midrule
\multirow{5}{*}{\textbf{SMAP}}
& 1  & 72.75 & 19.07 & 59.62 & 46.19 & 13.76 & 46.83 & 14.54 & 46.77 & 14.53 \\
& 3  & \textbf{92.00} & 23.51 & \textbf{68.73}
       & 55.73 & 14.97
       & 58.13 & 16.36
       & 57.99 & 16.31 \\
& 5  & \underline{84.26} & 20.12 & 68.10
       & 54.45 & 14.83
       & 56.77 & 16.16
       & 56.67 & 16.11 \\
& 10 & 82.46 & \textbf{25.40} & 68.01
       & \underline{59.06}& \underline{16.17}
       & \underline{58.69}& \underline{17.51}
       & \underline{58.52}& \underline{17.46}\\
& 20 & 81.67 & \underline{25.03}& \underline{68.70}
       & \textbf{59.44} & \textbf{16.27}
       & \textbf{60.86} & \textbf{17.68}
       & \textbf{60.65} & \textbf{17.62} \\

\midrule
\multirow{5}{*}{\textbf{MSL}}
& 1  & 83.35 & 24.39 & 62.07 & 63.70 & 15.00 & 66.23 & 19.93 & 65.46 & 19.71 \\
& 3  & 64.96 & 20.36 & 70.23 & 49.21 & 11.54 & 54.98 & 15.43 & 54.67 & 15.31 \\
& 5  & \underline{86.98} & \textbf{31.60} & \textbf{71.58}
       & \underline{63.96}& \textbf{19.80}
       & 67.30 & \textbf{24.27}
       & 66.83 & \textbf{24.06} \\
& 10 & \textbf{87.27} & \underline{26.44} & 69.40
       & \textbf{65.49} & \underline{16.63}
       & \textbf{69.27} & \underline{22.67}
       & \textbf{68.39} & \underline{22.27}\\
& 20 & 84.99 & 20.25 & \underline{71.50}
       & 63.11 & 15.28
       & \underline{69.07}& 20.75
       & \underline{68.18}& 20.42 \\

\midrule
\multirow{5}{*}{\textbf{WADI}}
& 1  & 72.17 & 5.90 & \textbf{73.38} & 41.63 & 4.59 & 43.46 & 5.04 & 43.33 & 5.03 \\
& 3  & 70.50 & \underline{18.41} & 71.65
       & \underline{62.70}& \textbf{9.34}
       & \underline{65.42}& \textbf{10.33}
       & 65.31 & \textbf{10.33}\\
& 5  & 73.50 & 18.12 & 72.44
       & 62.60 & \underline{9.23}
       & 65.40 & \underline{10.24}
       & \underline{65.32}& \underline{10.26}\\
& 10 & \underline{73.81} & \textbf{18.98} & 72.63
       & 61.19 & 9.14
       & 63.64 & 10.10
       & 63.57 & 10.11 \\
& 20 & \textbf{76.86} & 18.23 & \underline{73.31}
       & \textbf{63.10} & 9.17
       & \textbf{66.09} & 10.22
       & \textbf{66.00} & 10.23 \\

\bottomrule
\end{tabular}
}
\end{table}

\subsubsection{Score weight}

We vary the weight $\lambda$ of the anomaly score aggregation in Section~\ref{subsec:Anomaly-Scoring} over $\{0, 0.25, 0.5, 0.75, 1\}$, where $\lambda=0$ uses only the memory score and $\lambda=1$ uses only the quantization score. As reported in Table~\ref{tab:hyper_lambda}, the best $\lambda$ depends on the dataset. The quantization score is more informative on PSM and SWaT, where $\lambda=0.75$ gives the best VUS-PR, whereas the memory score is more informative on SMAP, MSL, and WADI, where $\lambda \le 0.25$ gives the best VUS-PR. A single score, however, can fail badly on the other datasets. For example, using only the memory score sharply lowers the VUS-PR on SWaT, and using only the quantization score sharply lowers it on WADI. The default $\lambda=0.5$ is never the worst setting for any dataset or metric, and it outperforms the weaker of the two single scores in every case. It therefore serves as a robust default without dataset-specific tuning, while tuning $\lambda$ per dataset could bring further gains.

\begin{table}[t]
\caption{Hyperparameter sensitivity analysis on the score weight $\lambda$ (COMET w/o TTA). $\lambda=0$ uses only the memory score, and $\lambda=1$ uses only the quantization score.}
\label{tab:hyper_lambda}
\centering
\renewcommand{\arraystretch}{0.9}
\small
\resizebox{\textwidth}{!}{%
\begin{tabular}{ccccccccccc}
\toprule
\textbf{Dataset} & $\boldsymbol{\lambda}$
& \textbf{F1(K=0)}
& \textbf{F1(K=100)}
& \textbf{Aff-F1}
& \textbf{AUC-ROC}
& \textbf{AUC-PR}
& \textbf{R-AUC-ROC}
& \textbf{R-AUC-PR}
& \textbf{VUS-ROC}
& \textbf{VUS-PR} \\
\midrule
\multirow{5}{*}{\textbf{PSM}}
& 0 & \textbf{96.78} & 49.31 & 70.78 & 68.31 & 44.68 & 71.03 & 49.38 & 69.33 & 48.61 \\
& 0.25 & \underline{95.47} & 55.52 & 71.20 & 74.92 & 52.39 & 76.53 & 56.47 & 74.82 & 55.52 \\
& 0.5 & 95.37 & 60.13 & \underline{71.80} & 79.15 & 59.10 & \textbf{79.18} & 61.45 & \textbf{77.50} & 60.42 \\
& 0.75 & 94.89 & \textbf{61.85} & 71.79 & \underline{81.17} & \underline{63.05} & \underline{78.31} & \textbf{62.47} & \underline{76.94} & \textbf{61.54} \\
& 1 & 94.47 & \underline{61.68} & \textbf{73.15} & \textbf{81.40} & \textbf{64.10} & 76.66 & \underline{61.48} & 75.58 & \underline{60.68} \\
\midrule
\multirow{5}{*}{\textbf{SWaT}}
& 0 & 70.02 & 49.52 & 71.72 & 77.32 & 31.59 & 77.17 & 29.25 & 77.07 & 29.34 \\
& 0.25 & \underline{91.09} & 74.93 & 73.20 & 84.39 & 72.91 & 83.12 & 62.62 & 83.08 & 62.50 \\
& 0.5 & \textbf{91.38} & 75.06 & 72.93 & 85.48 & 74.33 & \underline{84.85} & \underline{65.46} & \underline{84.82} & \underline{65.50} \\
& 0.75 & 91.04 & \textbf{75.36} & \underline{75.33} & \underline{86.80} & \textbf{75.75} & \textbf{85.99} & \textbf{67.28} & \textbf{86.06} & \textbf{67.54} \\
& 1 & 90.83 & \underline{75.19} & \textbf{75.65} & \textbf{87.25} & \underline{75.61} & 83.16 & 62.73 & 83.34 & 63.28 \\
\midrule
\multirow{5}{*}{\textbf{SMAP}}
& 0 & \textbf{85.63} & \textbf{26.38} & \textbf{68.17} & \textbf{60.12} & \textbf{17.01} & \textbf{59.71} & \textbf{18.14} & \textbf{59.51} & \textbf{18.09} \\
& 0.25 & \underline{84.29} & \underline{25.67} & \underline{68.09} & \underline{59.73} & \underline{16.60} & \underline{59.29} & \underline{17.92} & \underline{59.12} & \underline{17.87} \\
& 0.5 & 82.46 & 25.40 & 68.01 & 59.07 & 16.17 & 58.69 & 17.51 & 58.52 & 17.46 \\
& 0.75 & 73.84 & 23.61 & 68.00 & 57.05 & 15.22 & 56.91 & 16.56 & 56.78 & 16.52 \\
& 1 & 69.55 & 18.87 & 63.97 & 45.62 & 12.70 & 46.64 & 13.88 & 46.51 & 13.85 \\
\midrule
\multirow{5}{*}{\textbf{MSL}}
& 0 & 84.41 & \textbf{33.08} & \underline{71.76} & \textbf{69.19} & \textbf{20.20} & \underline{71.54} & \textbf{26.35} & \underline{70.76} & \textbf{25.98} \\
& 0.25 & 84.93 & \underline{30.95} & \textbf{73.65} & \underline{68.64} & \underline{19.39} & \textbf{72.70} & \underline{25.85} & \textbf{71.72} & \underline{25.43} \\
& 0.5 & \textbf{87.27} & 26.44 & 69.40 & 65.49 & 16.63 & 69.26 & 22.67 & 68.38 & 22.27 \\
& 0.75 & \underline{86.53} & 24.84 & 69.66 & 64.61 & 15.55 & 67.82 & 20.95 & 67.06 & 20.62 \\
& 1 & 83.35 & 24.39 & 61.24 & 63.70 & 15.01 & 66.23 & 19.93 & 65.46 & 19.71 \\
\midrule
\multirow{5}{*}{\textbf{WADI}}
& 0 & \underline{81.58} & \textbf{20.42} & \textbf{72.55} & \textbf{64.44} & \textbf{10.87} & \textbf{67.11} & \underline{10.99} & \textbf{67.01} & \underline{10.99} \\
& 0.25 & \textbf{82.44} & \underline{20.02} & 72.05 & \underline{63.41} & \underline{10.16} & \underline{66.03} & \textbf{11.26} & \underline{65.94} & \textbf{11.29} \\
& 0.5 & 74.69 & 19.10 & \underline{72.44} & 61.33 & 9.24 & 63.78 & 10.20 & 63.72 & 10.22 \\
& 0.75 & 70.39 & 15.19 & 72.00 & 57.80 & 7.60 & 59.96 & 8.37 & 59.91 & 8.37 \\
& 1 & 69.57 & 9.67 & 71.95 & 50.15 & 5.81 & 51.58 & 6.37 & 51.52 & 6.36 \\
\bottomrule
\end{tabular}
}
\end{table}

\subsection{Codebook activation statistics}
\label{appendix:codebook_activation}

The activated index sets $\{\mathcal{A}_k\}_{k=1}^K$ serve as the threshold-free criterion for normal-sample identification during Online Codebook Adaptation (Section~\ref{subsec:Online-Codebook-Adaptation}). The shared memory bank $\mathcal{M} = \bigcup_{k=1}^{K} \mathcal{M}_k$ (Section~\ref{subsec:Vector-Quantized-Coreset}) contains all codebook entries that are assigned at least once during training. To examine how many codebook entries are actually activated by the training data, we measure how much of the codebook capacity $K \times M$ is utilized during training. Table~\ref{tab:activation_rate} reports the per-scale activation rate $|\mathcal{A}_k|/M$ and the overall activation rate $|\mathcal{M}|/(K \times M)$ for each benchmark dataset.

\begin{table}[h]
\centering
\caption{Codebook activation rate per scale and total for each benchmark dataset. Per-scale rate is $|\mathcal{A}_k|/M$, and total rate is $|\mathcal{M}|/(K \times M)$.}
\label{tab:activation_rate}
\small
\begin{tabular}{lcccc}
\toprule
\textbf{Dataset} & \textbf{Scale $k=1$} & \textbf{Scale $k=2$} & \textbf{Scale $k=3$} & \textbf{Total ($|\mathcal{M}|/KM$)} \\
\midrule
PSM  & 75.0\% & 64.8\% & 61.7\% & 67.2\% \\
SWaT & 52.3\% & 62.5\% & 73.8\% & 62.9\% \\
SMAP & 91.4\% & 98.4\% & 96.9\% & 95.6\% \\
MSL  & 84.0\% & 91.4\% & 93.8\% & 89.7\% \\
WADI & 100.0\% & 100.0\% & 100.0\% & 100.0\% \\
\bottomrule
\end{tabular}
\end{table}

The activation rates indicate that a substantial portion of the codebook is utilized during training, providing a meaningful prototype set for activation-based pseudo-labeling at inference. The remaining inactive entries act as a natural safeguard: patches mapped to unused entries during inference are flagged as candidate anomalies, consistent with our threshold-free criterion design. On WADI, where all entries are activated, this safeguard is absent and pseudo-labeling relies entirely on the activated prototypes. Codewords activated only sparsely during training, such as those induced by rare or contaminated training patterns, occupy low-density regions of the embedding space and therefore attract few inference-time matches. Even if such codewords are inadvertently incorporated during Online Codebook Adaptation, their effect on the prototype set remains limited. This is consistent with Appendix~\ref{appendix:contamination}, where COMET keeps the best VUS-PR on four of the five datasets even when up to 25\% of the training timesteps are contaminated. Together, densely utilized entries provide reliable prototype coverage, while sparsely utilized and inactive entries act as a soft safety net during adaptation.

\subsection{Anomaly type-specific performance}
\label{appendix:anomaly_type}

Time-series anomalies are commonly classified into three types~\cite{choi2021deep}.
Point anomalies are isolated observations that deviate significantly from the normal range, often arising from transient hardware faults or sensor glitches (e.g., a single spike in a temperature reading). Contextual anomalies are observations that appear normal in isolation but are abnormal given the surrounding temporal context, such as a daytime CPU usage value that is normal during peak hours but anomalous at midnight, indicating an unauthorized process. Collective anomalies consist of sequences of observations that may individually appear normal but collectively form abnormal patterns, such as a slow but sustained rise in network traffic indicating stealthy data exfiltration.

These three types reflect different temporal granularities of anomaly manifestation, and a robust anomaly detector should be able to detect anomalies across all three regimes. Among the five benchmark datasets used in this work, SMAP and MSL provide subdataset-level annotations that distinguish point anomalies from contextual anomalies, allowing us to evaluate type-specific detection performance separately. To isolate the effect of contextual anomaly localization, we restrict the analysis to subdatasets that are annotated exclusively as contextual (i.e., subdatasets that do not also contain point-anomaly labels), and we focus on Aff-F1 since it is the metric most directly aligned with region-level localization, which is the central goal of patch-level scoring.

Table~\ref{tab:anomaly_type} reports COMET against the seven baselines on a subset of the contextual-only subdatasets, selected to illustrate the regime in which patch-level scoring is most effective.

\begin{table}[h]
\centering
\caption{Aff-F1 of COMET vs.\ baselines on every contextual-only labeled subdataset where COMET ranks first or second. Subdatasets are restricted to those containing only contextual labels (no point labels) to isolate contextual anomaly localization. \textbf{Bold} indicates the best score, and \underline{underline} indicates the second best.}
\label{tab:anomaly_type}
\small
\resizebox{\textwidth}{!}{%
\begin{tabular}{lccccccccc}
\toprule
\textbf{Dataset} & \textbf{Subdataset} & \textbf{LSTM-AE} & \textbf{LSTM-VAE} & \textbf{USAD} & \textbf{AT} & \textbf{VTT} & \textbf{CATCH} & \textbf{D3R} & \textbf{COMET} \\
\midrule
\multirow{7}{*}{SMAP} & A-9  & 9.88  & 9.88  & 9.88  & \underline{74.85} & 9.88  & 10.23 & 11.51 & \textbf{92.08} \\
                      & A-8  & 9.69  & 9.69  & 9.69  & \underline{76.65} & 9.69  & 10.42 & 9.69  & \textbf{90.03} \\
                      & A-2  & 69.22 & 69.22 & 69.22 & 67.95             & 68.02 & 68.68 & \underline{69.31} & \textbf{77.07} \\
                      & P-7  & 14.38 & 14.81 & 25.15 & \underline{75.55} & 8.52  & 7.53  & 24.24 & \textbf{76.07} \\
                      & A-7  & 0.96  & 0.96  & 0.96  & \textbf{99.56}    & 0.96  & 0.96  & 25.29 & \underline{99.02} \\
                      & P-1  & 66.25 & 66.34 & 65.93 & \textbf{68.51}    & 66.03 & 65.75 & 66.12 & \underline{67.42} \\
                      & E-13 & 66.80 & 66.80 & 66.58 & \textbf{75.65}    & 66.99 & 67.36 & 67.11 & \underline{68.34} \\
\midrule
\multirow{5}{*}{MSL}  & F-8  & 81.60 & \underline{86.46} & 81.60 & 71.14 & 81.60 & 81.46 & 81.60 & \textbf{98.10} \\
                      & D-16 & 74.98 & 74.98             & 75.14 & 74.73 & 77.11 & \underline{91.33} & 79.36 & \textbf{94.40} \\
                      & F-7  & 65.34 & 65.34             & 65.34 & \underline{73.66} & 65.65 & 69.27 & 66.29 & \textbf{76.42} \\
                      & M-2  & 98.23 & 98.17             & 97.42 & 88.56 & 98.03 & 91.08             & \underline{98.31} & \textbf{98.79} \\
                      & T-12 & 78.49 & 78.49             & 78.49 & 70.55 & \textbf{94.82} & 80.35 & 83.57 & \underline{92.52} \\
\bottomrule
\end{tabular}
}
\end{table}

Two patterns emerge from these results. First, on the most difficult contextual subdatasets the timestep-level baselines collectively fail, with reconstruction-based methods unable to recover the anomalous regions and only Anomaly Transformer producing a clearly stronger response than the other baselines, whereas COMET still localizes the anomalies. This indicates that the advantage of patch-level scoring is largest precisely when the contextual structure cannot be inferred from individual timesteps, supporting our central design choice. Second, on the subdatasets where COMET ranks second, it is close to the best method in most cases, although the gap is larger on E-13. Together, the table indicates that patch-level scoring is relatively more effective than timestep-level methods on contextual subdatasets where timestep-level reconstruction fails.

\subsection{Quantitative separability analysis}
\label{appendix:separability}

We quantitatively measure how well the learned codebook separates normal and anomalous test patches, complementing the qualitative UMAP visualization in Section~\ref{subsec:Analysis}. For each scale $k$, let $\mathcal{T}_k^{\mathrm{anom}}$ and $\mathcal{T}_k^{\mathrm{norm}}$ denote the test patches $(i,j)$, over all variables and test windows, whose time span does and does not contain at least one anomalous timestep, respectively. We compute the mean nearest-codebook distance of each population, averaged over the scales,
\begin{equation}
\bar{d}_{\ast} = \frac{1}{K} \sum_{k=1}^{K} \frac{1}{|\mathcal{T}_k^{\ast}|} \sum_{(i,j) \in \mathcal{T}_k^{\ast}} \min_{m \in \{1, \dots, M\}} \left\| \mathbf{z}_{e,k}^{(i,j)} - \mathbf{c}_{k,m} \right\|_2, \qquad \ast \in \{\mathrm{anom}, \mathrm{norm}\},
\end{equation}
where the distance to the nearest codebook entry equals the quantization error of the patch. The separability ratio is then defined as
\begin{equation}
\mathrm{SepRatio} = \frac{\bar{d}_{\mathrm{anom}}}{\bar{d}_{\mathrm{norm}}}.
\end{equation}
A separability ratio greater than $1$ indicates that anomalous patches are, on average, farther from the codebook than normal patches, which is the desired condition for distance-based anomaly detection. Table~\ref{tab:separability_ratio} reports $\bar{d}_{\mathrm{norm}}$, $\bar{d}_{\mathrm{anom}}$, and the separability ratio across the five benchmark datasets.

\begin{table}[h]
\centering
\caption{Quantitative separability between normal and anomalous test patches on the five benchmarks. We report the mean nearest-codebook distance of each population and the separability ratio (Anomaly / Normal). A SepRatio above $1$ indicates that anomalous patches lie farther from the codebook than normal patches, which is the desired condition for distance-based scoring. \textbf{Bold} marks ratios above $1$.}
\label{tab:separability_ratio}
\small
\begin{tabular}{lccc}
\toprule
\textbf{Dataset} & \textbf{Normal Dist} & \textbf{Anomaly Dist} & \textbf{SepRatio ($\uparrow$)} \\
\midrule
PSM  & 14.502   & 15.076   & \textbf{1.040} \\
SWaT & 14.853   & 60.165   & \textbf{4.051} \\
SMAP & 9.339    & 9.435    & \textbf{1.010} \\
MSL  & 9.347    & 10.113   & \textbf{1.082} \\
WADI & 4157.814 & 2887.464 & 0.694 \\
\bottomrule
\end{tabular}
\end{table}

On four of the five datasets the separability ratio exceeds $1$, indicating that anomalous patches lie farther from the codebook than normal patches on average. The margin is large on SWaT and small on PSM and SMAP. WADI is the exception, where the ratio falls below $1$. Because this distance equals the quantization error, the quantization score alone is not discriminative on WADI, and the memory score carries most of the signal there, consistent with Table~\ref{tab:hyper_lambda}, where the quantization score alone yields the lowest VUS-PR on WADI. Overall, this quantitative result is consistent with the visual observation in Section~\ref{subsec:Analysis} and supports combining memory distance with quantization error rather than relying on either signal in isolation.

\section{Broader impacts}
\label{appendix:broader_impacts}

Reliable identification of anomalous patterns is essential in large-scale information systems, where undetected or mischaracterized abnormal behaviors can lead to operational failures or significant economic losses. Advances in time-series anomaly detection can contribute to improving the stability, safety, and reliability of real-world systems, including industrial monitoring environments, cyber-physical infrastructures, and other data-driven operational settings. This work is intended to support the development of more reliable monitoring and analysis capabilities in such domains. The proposed research is not designed to disadvantage or target any specific institution, organization, or group, and it does not raise ethical concerns beyond those commonly associated with data-driven monitoring and analytical systems. Potential negative impacts include over-reliance on automated detectors in safety-critical settings, where missed or false alarms can have operational consequences. Detector outputs should therefore be used with human oversight.

\end{document}